\documentclass{article}

\usepackage{arxiv}

\usepackage[utf8]{inputenc}    
\usepackage[T1]{fontenc}       
\usepackage[colorlinks, linkcolor=black, anchorcolor=black, citecolor=blue]{hyperref} 
\usepackage{url}               
\usepackage{graphicx}          
\usepackage{subcaption}        
\usepackage{booktabs}          
\usepackage{amsmath}           
\usepackage{amssymb}           
\usepackage{amsfonts}          
\usepackage{nicefrac}          
\usepackage{microtype}         
\usepackage{lipsum}            
\usepackage{natbib}            
\usepackage{doi}               
\usepackage{multirow}          
\usepackage{array}             
\usepackage{wrapfig}           
\usepackage{caption}           
\usepackage{tabularx}          
\usepackage{booktabs}          

\usepackage{enumitem}          
\usepackage{float}             
\usepackage{makecell}          
\usepackage{bm}                
\usepackage{algorithm}
\usepackage{algorithmic}
\title{Adaptive Pruning with Module Robustness Sensitivity: Balancing Compression and Robustness}

\date{} 					

\author{
{\large  Lincen BAI$^{1}$ ~ Hedi Tabia$^{1}$ ~ Raúl Santos-Rodríguez$^{2}$} \\
\\
{\large $^{1}$ Université Paris-Saclay   ${^2}$ University of Bristol} 
}

\renewcommand{\headeright}{}
\renewcommand{\undertitle}{}
\renewcommand{\shorttitle}{}

\hypersetup{
pdftitle={Beyond Pruning Criteria: The Dominant Role of Fine-Tuning and Adaptive Ratios in Neural Network Robustness},
pdfsubject={q-bio.NC, q-bio.QM},
pdfauthor={Lincen BAI.~Hedi Tabia, Raúl Santos-Rodríguez},
}

\begin{document}
\maketitle

\begin{abstract}
Neural network pruning has traditionally focused on weight-based criteria to achieve model compression, frequently overlooking the crucial balance between adversarial robustness and accuracy. Existing approaches often fail to preserve robustness in pruned networks, leaving them more susceptible to adversarial attacks. This paper introduces Module Robustness Sensitivity (MRS), a novel metric that quantifies layer-wise sensitivity to adversarial perturbations and dynamically informs pruning decisions. Leveraging MRS, we propose Module Robust Pruning and Fine-Tuning (MRPF), an adaptive pruning algorithm compatible with any adversarial training method, offering both flexibility and scalability. Extensive experiments on SVHN, CIFAR, and Tiny-ImageNet across diverse architectures, including ResNet, VGG, and MobileViT, demonstrate that MRPF significantly enhances adversarial robustness while maintaining competitive accuracy and computational efficiency. Furthermore, MRPF consistently outperforms state-of-the-art structured pruning methods in balancing robustness, accuracy, and compression. This work establishes a practical and generalizable framework for robust pruning, addressing the long-standing trade-off between model compression and robustness preservation.
\end{abstract}


\section{Introduction}

Deep neural networks (DNNs) have driven remarkable advancements in fields such as image recognition~\citep{simonyan2014very} and natural language processing~\citep{kenton2019bert}, achieving state-of-the-art performance across various applications. However, the growing size and complexity of DNNs present significant challenges for deployment, especially in resource-constrained environments where computational and energy efficiency are critical. To address these challenges, model compression techniques, such as pruning~\citep{han2015learning, he2019filter}, quantization~\citep{wu2016quantized}, and knowledge distillation~\citep{hinton2015distilling}, have been widely adopted for reducing model size while maintaining acceptable performance. Among these, pruning has become a popular choice due to its simplicity and ability to eliminate redundant parameters.

Pruning, while effective in reducing model size and computational demands, introduces significant challenges in adversarial scenarios. By removing parameters, pruning often compromises the decision boundaries essential for accurate predictions, making sparse models—common outcomes of aggressive pruning—more vulnerable to adversarial attacks. Such attacks exploit the weakened decision boundaries, leading to severe performance degradation~\citep{sun2023simple}. Furthermore, there is strong evidence indicating that achieving adversarial robustness often requires over-parameterized networks with wider layers and increased structural complexity, inherently conflicting with the objectives of model compression~\citep{madry2017towards, zhang2019theoretically, wu2021wider}. For example,~\citet{madry2017towards} demonstrate that the robust decision boundaries required for adversarially trained networks are far more complex than those needed for standard data classification, requiring larger model capacities to withstand strong adversarial attacks. Similarly,~\citet{zhang2019theoretically} characterize the trade-off between robustness and accuracy, revealing that the robust error depends on both natural error and the complexity of the decision boundary.

\begin{figure}[H]
    \centering
    \begin{subfigure}{0.48\linewidth}
        \centering 
        \includegraphics[width=\linewidth]{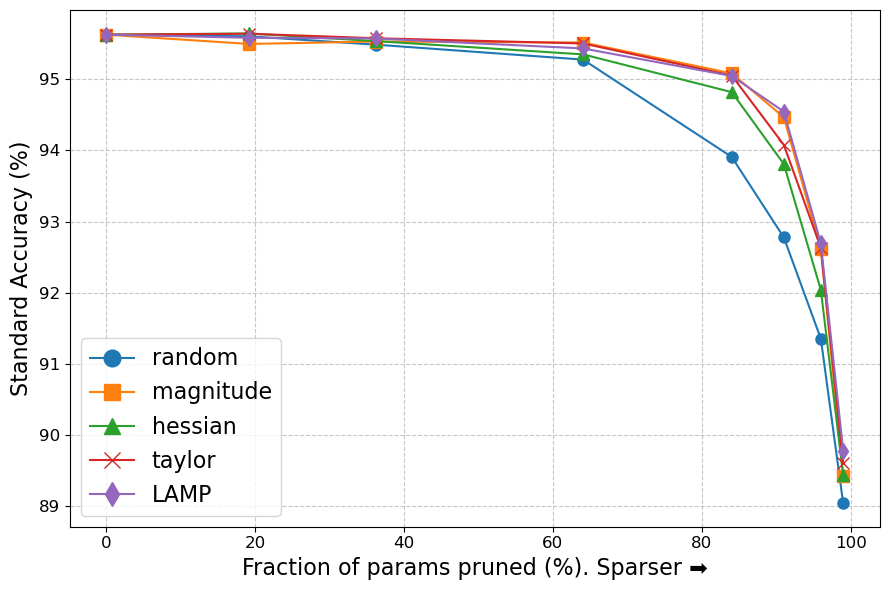}
        \captionsetup{font=scriptsize} 
        \caption*{} 
        \label{fig:resnet50_acc_pruning}
    \end{subfigure}
    \hfill
    \begin{subfigure}{0.48\linewidth}
        \centering
        \includegraphics[width=\linewidth]{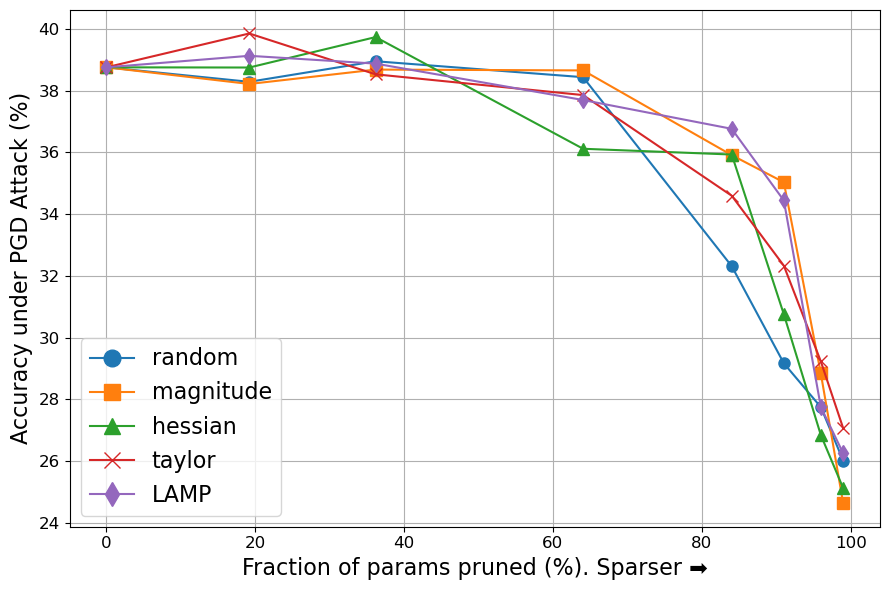}
        \captionsetup{font=scriptsize} 
        \caption*{}
        \label{fig:resnet50_rob_pruning}
    \end{subfigure}
    
    \caption{
        The impact of different pruning criteria (random, magnitude, Hessian, Taylor, LAMP) on accuracy and robustness for ResNet-50 on CIFAR-10. Both accuracy and robustness decline as the model becomes sparser, especially at high compression ratios. The choice of pruning criterion has limited influence on the final performance.
    }
    \label{fig:pruning_impact_resnet50}
\end{figure}
These challenges are further compounded by the findings that wider networks, while improving natural accuracy, often exhibit lower perturbation stability, adversely affecting overall robustness unless robust regularization parameters are carefully optimized~\citep{wu2021wider}. This trade-off highlights the inherent difficulty of reconciling model compression with robustness preservation, as robust models require sufficient capacity to maintain stable decision boundaries under adversarial perturbations. Consequently, designing pruning strategies that achieve both computational efficiency and adversarial robustness remains an open and critical problem in neural network optimization.

Through extensive experiments, we observe a key limitation in traditional pruning approaches: while pruning criteria, such as weight magnitude or importance scores, guide the parameter removal process, they have minimal impact on the ultimate performance of pruned networks after fine-tuning. Instead, as shown in \autoref{fig:pruning_impact_resnet50}, both accuracy and robustness decline significantly as the pruning ratio increases, especially at higher compression rates. This trend is consistent across multiple pruning criteria, highlighting that fine-tuning—particularly adversarial fine-tuning—plays a dominant role in restoring performance after pruning.

Motivated by these findings, we propose \textbf{Module Robust Pruning and Fine-Tuning (MRPF)}, a novel framework that bridges the gap between model compression and adversarial robustness. At the core of MRPF is the \textbf{Module Robustness Sensitivity (MRS)} metric, which measures the sensitivity of individual layers to adversarial perturbations. By dynamically adjusting pruning ratios based on MRS, MRPF ensures the retention of robustness-critical parameters, enabling both efficiency and resilience against adversarial attacks. Importantly, MRPF is compatible with existing adversarial training methods, offering flexibility and scalability.

Our contributions are summarized as follows:
\begin{itemize}
    \item We identify and empirically validate that the role of pruning criteria in determining the final performance of pruned networks is limited, highlighting the dominant importance of fine-tuning in restoring adversarial robustness.
    \item We propose \textbf{Module Robustness Sensitivity (MRS)}, a novel metric for layer-wise sensitivity analysis, and integrate it into the \textbf{MRPF} framework to adaptively prune networks while preserving robustness-critical parameters.
    \item Extensive experiments on SVHN, CIFAR-10/100, and Tiny-ImageNet across diverse architectures, including ResNet, VGG, and MobileViT, demonstrate that MRPF achieves significant improvements in adversarial robustness while maintaining competitive accuracy and efficiency.
\end{itemize}

\section{Related Work}

\subsection{Pruning Neural Networks} 
Pruning research dates back to 1988~\citep{hanson1988comparing}, but it gained widespread attention when ~\citet{han2015learning} proposed sparsifying deep models by removing non-significant synapses and then re-training to restore performance. The Pruning After Training (PAT) pipeline is the most popular because it is believed that pretraining a dense network is essential to obtain an efficient subnetwork~\citep{frankle2018lottery}.

Weight importance can be assessed through various metrics, such as the magnitude of the weights. ~\citet{han2015learning} evaluated weight importance by absolute value, while ~\citet{li2016pruning} used the l1-norm to score filters. More advanced techniques, such as Discrimination-aware Channel Pruning (DCP)~\citep{zhuang2018discrimination}, incorporate loss minimization to identify the most discriminative channels.

The Lottery Ticket Hypothesis (LTH)~\citep{frankle2018lottery} posits that sparse, independently trainable subnetworks can be found within dense networks. However, later works, including Multi-Prize LTH~\citep{diffenderfer2021multi}, challenge LTH, showing that training settings like learning rate and epochs greatly influence the success of finding such subnetworks.

Recent efforts aim to achieve more precise pruning by leveraging deeper insights into network structure and function. For instance, Group Fisher Pruning (GFP)~\citep{liu2021group} evaluates importance by computing the Fisher information, providing a statistical measure of each parameter's contribution to the network's performance. Similarly, techniques like HRank~\citep{lin2020hrank} use Singular Value Decomposition to rank filters based on the information content of their activations, effectively identifying and removing less informative filters.

\subsection{Robustness of Neural Networks}
The susceptibility of state-of-the-art DNNs to adversarial attacks was first highlighted by \citet{goodfellow2014explaining} and \citet{szegedy2013intriguing}. Adversarial attacks are subtle manipulations, often undetectable to humans, but can significantly affect model predictions. This vulnerability is particularly concerning in critical applications like autonomous driving and healthcare.

While pruning generally aims to improve efficiency, it often comes at the cost of reduced adversarial robustness~\citep{gui2019model}. However, studies such as~\citep{guo2018sparse, jordao2021effect} suggest that retraining after pruning can mitigate some of these negative effects. \citet{Phan2022CSTARTC} introduces CSTAR, a framework that integrates low-rankness and adversarial robustness into a unified optimization problem. It proposes an automatic rank selection scheme to avoid manual tuning and demonstrates significant improvements in benign and robust accuracy. Similarly, Adversarial Weight Perturbation (AWP)~\citep{wu2020adversarial} enhances robustness by applying adversarial perturbations to model weights during training, complementing pruning-based methods by targeting robustness directly.

\subsection{Robust Pruning} 
Recent advancements in pruning and fine-tuning strategies have emphasized improving both model efficiency and robustness. \citet{zhu2023improving} introduced Module Robust Criticality (MRC) to identify redundant modules in adversarially trained models, paired with Robust Critical Fine-Tuning (RiFT) to enhance generalization while preserving robustness by selectively fine-tuning robustness-critical modules. Similarly, \citet{bair2023adaptive} proposed Adaptive Sharpness-Aware Pruning (AdaSAP), which improves accuracy and robust generalization through adaptive weight perturbations and structured pruning, highlighting the role of layer-wise sensitivity during pruning. HARP \cite{zhao2024holistic} integrates global objective optimization with adversarial training to implement hardware-aware pruning strategies, using non-uniform layer compression rates to enhance robustness and efficiency simultaneously. BCS \cite{ozdenizci2021training} leverages probabilistic sampling methods to train sparse networks robustly by exploring and preserving optimal connectivity patterns, ensuring robust generalization under extreme sparsity.

In comparison, MRPF uniquely employs MRS-guided dynamic pruning to prioritize robustness-critical layers and explicitly integrates adversarial fine-tuning post-pruning, offering a layer-specific and fine-grained approach. Unlike methods such as ICDM \cite{jian2022pruning}, which avoids adversarial training during pruning by relying on pre-trained robust networks, or HYDRA \cite{sehwag2020hydra}, which applies global pruning objectives, MRPF demonstrates superior adaptability across datasets and architectures, achieving a refined balance between robustness, accuracy, and compression, even under challenging sparsity conditions.

\subsection{Fine-tuning after Pruning}
Fine-tuning is a crucial step in recovering performance after pruning~\citep{le2021network}. It can be performed either by fine-tuning the pruned model or training it from scratch, and the choice between these approaches has been extensively debated~\citep{cheng2024survey}. Research by \citet{liu2018rethinking} suggests that subnetworks trained from random reinitialization often outperform those fine-tuned from pre-trained models, particularly in architectures like ResNet and VGG on ImageNet. However, \citet{li2016pruning} demonstrate that fine-tuning yields better results than training from scratch in their experiments. Studies by \citet{gao2021network} emphasize that fine-tuning is critical for achieving superior performance in sparse mobile networks, outperforming models trained from scratch.

In adversarial settings, fine-tuning after pruning becomes even more critical. Common adversarial training (AT) methods, such as PGD~\citep{madry2017towards}, TRADES~\citep{zhang2019theoretically}, and MART~\citep{wang2019improving}, have proven effective in improving robustness. PGD-based training directly minimizes the worst-case loss within a bounded perturbation region, ensuring robustness against strong attacks. TRADES formalizes a trade-off between robustness and accuracy by balancing natural and adversarial losses, while MART emphasizes misclassified examples during adversarial training, thereby focusing on difficult decision boundaries. Integrating these AT methods during fine-tuning helps restore both accuracy and robustness in pruned networks. 

By incorporating adversarial fine-tuning, pruning strategies can better handle the trade-off between efficiency and robustness, as demonstrated in recent works like CSTAR~\citep{Phan2022CSTARTC}. Our proposed MRPF framework leverages such adversarial fine-tuning techniques while introducing Module Robustness Sensitivity (MRS) to dynamically adjust pruning ratios, ensuring robust decision boundaries are preserved even in sparse models.

\section{Proposed Method} \label{sec:method}

\begin{figure*}[t]
    \centering
    \includegraphics[width=0.98\textwidth]{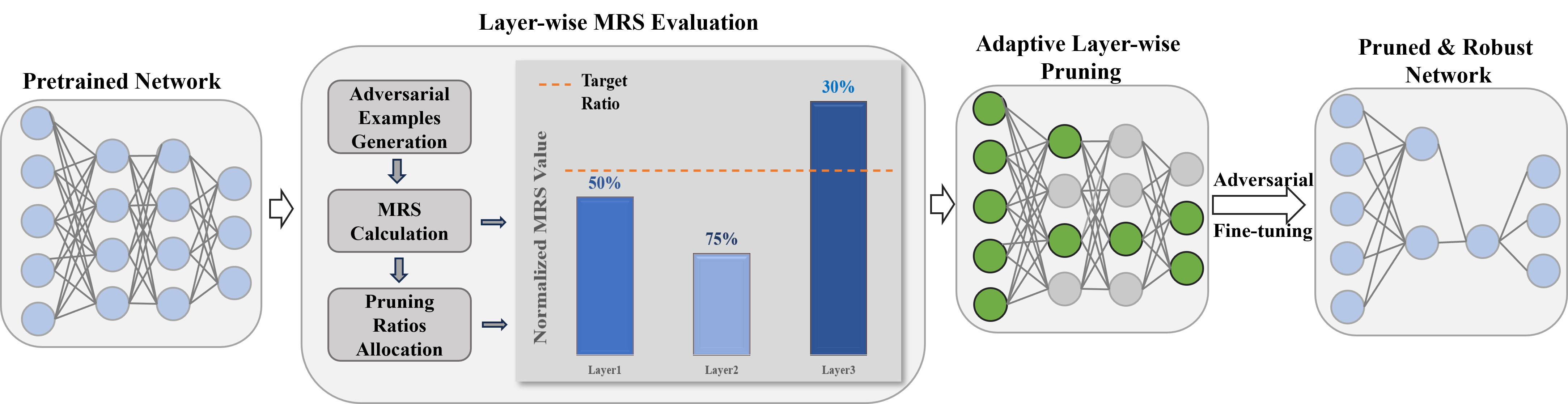} 
    \captionsetup{font=scriptsize} 
    \caption{
        Illustration of the proposed Module Robust Pruning and Fine-Tuning (MRPF) method. The process starts with a pre-trained network, generates adversarial examples, calculates layer-wise Module Robustness Sensitivity (MRS), and allocates pruning ratios dynamically. Critical layers are retained during pruning(layers with lower MRS values are pruned more aggressively), and adversarial fine-tuning ensures the pruned network maintains both robustness and accuracy.
    }
    \label{fig:mrpf_pipeline}
\end{figure*}

Our proposed method, Module Robust Pruning and Fine-Tuning (MRPF), as demonstrated in Figure~\ref{fig:mrpf_pipeline}, directly addresses the trade-off between model compression and robustness by incorporating adversarial examples into the pruning process.The core insight underlying MRPF is that robustness is not evenly distributed across layers in a neural network. Certain layers play a more critical role in maintaining adversarial resilience, while others contribute less. By dynamically calculating layer-wise Module Robustness Sensitivity (MRS) and retaining robustness-critical parameters during pruning, MRPF achieves a highly compressed model with preserved security.

The rationale for this approach is two-fold. First, conventional pruning methods often prioritize compression metrics, neglecting the network's robustness. This oversight leads to compromised defenses when exposed to adversarial examples. By integrating adversarial sensitivity into the pruning process, MRPF ensures the retention of critical parameters that uphold robustness, even in a more compact form. Second, MRPF is computationally efficient. By focusing pruning decisions on the sensitivity of individual layers, the method optimizes resource allocation, targeting pruning where it has the least impact on robustness. This targeted approach not only reduces FLOPs but also enhances the network's efficiency without sacrificing much adversarial robustness.

\subsection{Preliminaries}

We define a neural network with \( L \) layers, where \( N_i \) and \( N_{i+1} \) represent the number of input and output channels of the \( i^{th} \) convolutional layer, respectively. Let \( K \) denote the kernel size, and \( F_{i,j} \in \mathbb{R}^{N_i \times K \times K} \) represent the \( j^{th} \) filter in the \( i^{th} \) layer. The complete set of filters for the \( i^{th} \) layer is denoted by \( \{F_{i,j}\}_{j=1}^{N_{i+1}} \) and is parameterized by weights \( W(i) \in \mathbb{R}^{N_{i+1} \times N_i \times K \times K} \) for all layers \( i \in [1, L] \).

For pruning, we partition the filters into two subsets: \( F_{pruned} \) for removal and \( F_{keep} \) for retention. \( F_{keep} \) is defined as the set \( \{ F_{i,j} | j \in ID(i) \} \), where \( ID(i) \) indexes the most critical filters contributing significantly to performance in layer \( i \).

Given a dataset \( D = \{(x_i, y_i)\}_{i=1}^n \) and a desired sparsity level \( \kappa \) (representing the number of remaining non-zero filters), the objective is to minimize the loss function \( \ell(F_{keep}; D) \) over the set of retained filters \( F_{keep} \), subject to the constraint that the cardinality of \( F_{keep} \), denoted \( N_{set}(F_{keep}) \), does not exceed \( \kappa \):

\begin{equation}
\begin{split}
\min_{F_{keep}} \ell(F_{keep}; D) = \min_{F_{keep}} \frac{1}{n} \sum_{i=1}^{n} \ell(F_{keep}; (x_i, y_i)) \\
\text{s.t.} \quad N_{set}(F_{keep}) \leq \kappa, \quad F \in \mathbb{R}^{N \times K \times K}.
\end{split}
\end{equation}

The loss function \( \ell(\cdot) \) is typically cross-entropy loss. The reduction in FLOPs is measured after each iteration using:

\begin{equation}
P' = 1 - \frac{f_{flop}(N')}{f_{flop}(N)}
\end{equation}

This indicates the efficiency of pruning compared to the initial target.

\subsection{Pruning Method}
As illustrated in Figure~\ref{fig:mrpf_pipeline}, this approach ensures that the model maintains its defense against adversarial attacks even after significant pruning. The details are outlined below.
\textbf{Generate Adversarial Examples:} Adversarial examples \( AE \) are generated using an adversarial attack method \( AT \) on the training data \( D \). These examples are created to exploit the vulnerabilities of the model.

\begin{equation} \label{eq:adv_examples}
AE = AT(D)
\end{equation}

The baseline adversarial loss \( L_{\text{orig}} \) is computed by evaluating the adversarial examples on the original model:

\begin{equation} \label{eq:baseline_loss}
L_{\text{orig}} = L_{\text{adv}}(N(AE), y_{AE})
\end{equation}

This step is important because traditional pruning methods rely solely on performance metrics like accuracy. However, in adversarial settings, it is not enough for a pruned model to perform well on clean data—it must also withstand adversarial attacks. By incorporating adversarial examples into the evaluation, we ensure the pruned model is robust across a wider range of inputs.

\textbf{Perturb Layer Weights:} For each layer \( l_i \), the algorithm perturbs the weights \( W_{l_i} \) by performing gradient ascent on the adversarial loss. This step measures how sensitive the adversarial loss is to changes in the layer's parameters. The update rule for the weights is:

\begin{equation} \label{eq:weight_update}
W_{l_i}^{(t+1)} = W_{l_i}^{(t)} + \eta \cdot \nabla_{W_{l_i}} L_{\text{adv}}(N(W_{l_i}^{(t)}, X_{\text{adv}}), y_{\text{adv}})
\end{equation}

where \( \eta \) is the learning rate. After each update, the weights are projected back into a bounded region to satisfy the perturbation constraint:

\begin{equation} \label{eq:perturbation_constraint}
\| W_{l_i}^{(t+1)} - W_{l_i}^{\text{orig}} \|_2 \leq \epsilon \cdot \| W_{l_i}^{\text{orig}} \|_2
\end{equation}

Layers that are more sensitive to these perturbations play a larger role in the model's defense and should be pruned with caution. This step is justified because it allows us to identify the layers that contribute most to adversarial robustness, guiding the pruning process based on the model’s defensive properties rather than arbitrary weight magnitudes.

\textbf{Calculate Module Robustness Sensitivity (MRS):} 
We define the MRS to quantify the effect of perturbing each layer's weights on the adversarial loss. It is computed as the difference between the perturbed adversarial loss and the original adversarial loss:

\begin{equation} \label{eq:mrs_calculation}
\begin{split}
\text{MRS}(l_i) = L_{\text{adv}}(N(W_{l_i}^{\text{pert}}, AE), y_{AE}) \\
- L_{\text{adv}}(N(W_{l_i}^{\text{orig}}, AE), y_{AE})
\end{split}
\end{equation}
Layers with higher MRS values are more important for maintaining robustness, and their channels should be retained during pruning. The rationale for using MRS is that it directly ties the pruning decision to the network's robustness, ensuring that the layers critical for defense are preserved.

\textbf{Compute Pruning Ratios:} 
The pruning ratios are computed based on the inverse of the MRS values. Layers with lower MRS values (i.e., less critical for robustness) are pruned more aggressively, while layers with higher MRS values are pruned conservatively. This method ensures that pruning is guided by the functional importance of each layer rather than uniformity across the network, optimizing for both model compression and robustness.

The inverse of the MRS values is calculated:

\begin{equation} \label{eq:inv_mrs}
\text{InvMRS}(l_i) = \frac{1}{\text{MRS}(l_i) + \delta}
\end{equation}

where \( \delta \) is a small constant to prevent division by zero. The pruning weights are normalized across all layers:

\begin{equation} 
w_i = \frac{\text{InvMRS}(l_i)}{\sum_j \text{InvMRS}(l_j)}
\end{equation}

The pruning ratio for each layer is scaled by the global pruning ratio \( r_g \), capped at the maximum pruning ratio \( r_{\text{max}} \):

\begin{equation} \label{eq:prune_ratio}
p_i = \min(w_i \cdot r_g, \, r_{\text{max}})
\end{equation}

\textbf{Prune the Network:} The algorithm prunes each layer by removing the lowest \( p_i \times k_i \) channels based on importance scores (e.g., Taylor~\citep{molchanov2019importance}), where \( k_i \) is the number of channels in layer \( l_i \). The number of pruned channels is:

\begin{equation} \label{eq:channels_pruned}
k_i^{\text{pruned}} = \lfloor p_i \times k_i \rfloor
\end{equation}

\textbf{Fine-tune the Pruned Model:} After pruning, the model \( N' \) is fine-tuned using an augmented dataset \( D' \) that includes a fraction \( r_{AT} \) of newly generated adversarial examples:

\begin{equation}
D' = D \cup \left( r_{AT} \times AE' \right)
\end{equation}

For fine-tuning, we employ adversarial training using the TRADES framework~\citep{zhang2019theoretically}, which balances natural accuracy and robustness by minimizing a trade-off between the standard cross-entropy loss and an adversarial robustness loss. The TRADES loss function is defined as:

\begin{equation}
L_{\text{TRADES}} = L_{\text{CE}}(N'(x), y) + \lambda \cdot \max_{\|x' - x\| \leq \epsilon} \text{KL}(N'(x) \| N'(x'))
\end{equation}

Here, \( L_{\text{CE}} \) is the standard cross-entropy loss, \( \text{KL}(\cdot \| \cdot) \) is the Kullback-Leibler divergence, \( \lambda \) is a weighting parameter controlling the trade-off, and \( \epsilon \) represents the perturbation bound. The adversarial examples \( x' \) are generated iteratively using projected gradient descent (PGD). 

While TRADES is used as the default adversarial training method, alternative approaches such as MART~\citep{wang2019improving} and PGD-based adversarial training~\citep{madry2017towards} can also be applied. 

The proposed algorithm (see \autoref{appendix_algorithm}), integrated into the Module Robust Pruning and Fine-tuning (MRPF) framework, combines three key criteria—gradient information, loss change after channel removal, and sensitivity to adversarial perturbations—to perform robust channel pruning. By generating adversarial examples and calculating the Module Robustness Sensitivity (MRS), the algorithm dynamically adjusts per-layer pruning ratios based on each layer’s contribution to robustness. This ensures that channels are pruned not only by weight magnitude but also by their functional importance to adversarial resilience. Fine-tuning with adversarial examples restores both accuracy and robustness, making it highly effective for deploying robust and efficient networks.

\section{Experiments}

In this section, we detail the experimental setup and the results obtained from our model pruning experiments across various architectures and datasets. Our goal is to explore the effects of different pruning strategies on both standard accuracy and adversarial robustness.

We begin by introducing the datasets and baseline architectures utilized for evaluation, followed by the training configurations tailored for pruning. Finally, the results are presented in terms of key metrics such as test accuracy and adversarial robustness. More information regarding the experiments is provided in Appendix \ref{appendix_exp}.

\subsection{Experimental Setup}

\paragraph{Datasets and Baseline Architectures.}
We conduct experiments on four widely-used datasets: \textbf{CIFAR-10}, \textbf{CIFAR-100}~\citep{Krizhevsky09},\textbf{SVHN}~\citep{netzer2011reading} , and \textbf{Tiny-ImageNet}~\citep{le2015tiny}. These datasets offer diverse complexity levels, enabling a robust evaluation of pruning methods under varying conditions. Our experiments leverage popular architectures, including \textbf{VGGNet}~\citep{simonyan2014very}, \textbf{ResNet}~\citep{he2016identity}, and \textbf{MobileViT}~\citep{mehta2021mobilevit}. Each architecture is pruned with channel compression ratios ranging from 10\% to 80\%, simulating varying levels of sparsity.

\paragraph{Training Configurations.}
The training setup closely follows prior works~\citep{he2016identity,mehta2021mobilevit,he2019filter}, tailored specifically for pruning experiments. Models are fine-tuned using SGD with momentum, and two learning rate schedules are employed: \textbf{Step Annealing}, which reduces the learning rate to 10\% of its previous value at fixed intervals (e.g., every 10 epochs), and \textbf{Cosine Annealing}, which decays the learning rate smoothly following a cosine function.

For fine-tuning after pruning, we adopt advanced adversarial training frameworks such as \textbf{TRADES}~\citep{zhang2019theoretically}, \textbf{MART}~\citep{wang2019improving}, and \textbf{PGD-based adversarial training}~\citep{madry2017towards}, which are widely recognized for balancing natural accuracy and adversarial robustness. Additionally, we incorporate a simplified adversarial training (AT) strategy by including a small ratio (e.g., 20\%) of perturbed examples in the training process to efficiently evaluate the effectiveness of our proposed method.

Fine-tuning settings for each method adhere to the standard configurations provided in their respective works. For complete hyperparameter settings and fine-tuning configurations are available in \autoref{sec:hyperparams_training}.

\paragraph{Evaluation Metrics.}
We evaluate the pruned models on:
\begin{itemize}
    \item \textbf{Standard Accuracy(SAcc):} Test set accuracy on CIFAR-10, CIFAR-100, Tiny-ImageNet, and SVHN.
    \item \textbf{Accuracy under Adversarial Attack(Adv):} Robust accuracy of models against different adversarial attacks:
    \begin{itemize}
        \item \textbf{PGD}~\citep{madry2017towards}: \( \epsilon = 8/255 \), step size \( 2/255 \).
        \item \textbf{FGSM}~\citep{goodfellow2014explaining}: Single-step attack with \( \epsilon = 8/255 \).
        \item \textbf{AutoAttack(AA)}~\citep{croce2020reliable}: An ensemble of adversarial methods, \( \epsilon = 8/255 \).
    \end{itemize}
\end{itemize}

All experiments are repeated three times with different random seeds, and we report the mean accuracy and robustness metrics to ensure reliability. 

\subsection{Main Results}
\label{sec:main_res}
\begin{table*}[t!]
\caption{The evaluation results for structured pruning methods on CIFAR10. SAcc and accuracy under AA  are reported for sparsity rates, with ResNet18 (52.56\%, 74.47\%, 84.42\%) and VGG16 (58.30\%, 79.83\%, 88.71\%) sparsity levels shown in the figures. Bold numbers indicate the best performance in each setting.
}
\label{tab:eval_S_cifar}
\centering
\resizebox{0.775\textwidth}{!}{%
\begin{tabular}{@{}ccccccc@{}}
\toprule
 &
  \multicolumn{6}{c}{\textbf{Structured Pruning CIFAR10 - SAcc(\%)/Adv$_{\text{AA}}$(\%)}} \\ \cmidrule(l){2-7} 
 &
  \multicolumn{3}{c|}{\textbf{ResNet18}} &
  \multicolumn{3}{c}{\textbf{VGG16}} \\ 
\multirow{-3}{*}{\textbf{Name}} &
  50 &
  75 &
  \multicolumn{1}{c|}{90} &
  50 &
  75 &
  90 \\ \midrule
RADMM &
  79.27/42.68 &
  78.81/40.79 &
  \multicolumn{1}{c|}{70.53/37.30} &
  74.58/39.67 &
  70.51/37.74 &
  58.58/31.79 \\
HARP &
  77.38/42.73 &
  80.06/42.09 &
  \multicolumn{1}{c|}{77.88/41.59} &
  76.70/40.01 &
  73.61/39.14 &
  66.45/35.62 \\
PwoA &
  83.44/44.79 &
  81.77/37.85 &
  \multicolumn{1}{c|}{76.41/28.56} &
  66.33/30.15 &
  63.36/24.91 &
  57.71/18.39 \\
TwinRep &
  79.90/45.58 &
  79.37/45.21 &
  \multicolumn{1}{c|}{78.41/44.30} &
  77.65/43.13 &
  77.58/42.77 &
  76.26/42.14 \\ \midrule
Ours &
  \textbf{83.58/48.78} &
  \textbf{81.89/50.86} &
  \multicolumn{1}{c|}{\textbf{81.15/46.44}} &
  \textbf{84.52/55.39} &
  \textbf{84.46/51.18} &
  \textbf{83.42/48.32} \\
\bottomrule
\end{tabular}%
}
\end{table*}

\begin{table*}[t!]
\caption{The evaluation results for structured pruning methods on SVHN. SAcc and accuracy under AA are reported for sparsity rates, with ResNet18 (51.32\%, 76.88\%, 94.02\%) and VGG16 (53.27\%, 78.59\%, 83.41\%) sparsity levels shown in the figures. Bold numbers indicate the best performance in each setting.}
\label{tab:eval_S_svhn}
\centering
\resizebox{0.775\textwidth}{!}{%
\begin{tabular}{@{}ccccccc@{}}
\toprule
 &
  \multicolumn{6}{c}{\textbf{Structured Pruning SVHN - SAcc(\%)/Adv$_{\text{AA}}$(\%)}} \\ \cmidrule(l){2-7} 
 &
  \multicolumn{3}{c|}{\textbf{ResNet18}} &
  \multicolumn{3}{c}{\textbf{VGG16}} \\ 
\multirow{-3}{*}{\textbf{Name}} &
  50 &
  75 &
  \multicolumn{1}{c|}{90} &
  50 &
  75 &
  90 \\ \midrule

HARP &
  91.72/45.82 &
  \textbf{92.07/46.80} &
  \multicolumn{1}{c|}{91.03/45.25} &
  91.53/44.10 &
  89.06/42.45 &
  87.89/39.25 \\
PwoA &
  92.56/41.68 &
  92.61/38.69 &
  \multicolumn{1}{c|}{91.42/31.69} &
  89.16/39.09 &
  89.22/33.89 &
  87.17/24.55 \\
TwinRep &
  90.71/37.33 &
  88.71/45.28 &
  \multicolumn{1}{c|}{85.44/45.10} &
  89.91/45.82 &
  87.10/43.26 &
  89.61/44.83 \\ \midrule
Ours &
  \textbf{91.77/46.92} &
  91.62/46.21 &
  \multicolumn{1}{c|}{\textbf{90.95/47.07}} &
  \textbf{91.45/47.58} &
  \textbf{89.50/47.33} &
  \textbf{89.86/46.44} \\
\bottomrule
\end{tabular}%
}
\end{table*}

In this section, we present a comparative evaluation of our method against several state-of-the-art structured pruning approaches, including Robust-ADMM~\citep{ye2019adversarial}, HARP~\citep{zhao2024holistic}, PwoA~\citep{jian2022pruning}, and TwinRep~\citep{li2023learning}. The results leverage the standardized re-evaluation framework provided by \citep{piras2024adversarial}, which establishes consistent and credible baselines for structured pruning methods across CIFAR-10 and SVHN. For consistency, we adopt approximate parameter sparsity levels of 50\%, 75\%, and 90\%. These systematic comparisons validate our theoretical framework's ability to preserve adversarial robustness without compromising compression efficacy. 

\paragraph{Performance on CIFAR-10}
The CIFAR-10 results validate the robustness and effectiveness of our method, especially under extreme sparsity conditions. At 90\% sparsity, our approach achieves \textbf{81.15\%} SAcc and \textbf{46.44\%} AA robustness for ResNet-18, outperforming TwinRep by +2.14\% in robustness while methods like PwoA experience significant degradation, dropping to just 28.56\%. Similarly, for VGG-16, our method achieves \textbf{83.42\%} SAcc and \textbf{48.32\%} robustness at 90\% sparsity, marking a +7.16\% improvement in robustness compared to TwinRep. 

\paragraph{Performance on SVHN}
On SVHN, our method continues to demonstrate strong performance, maintaining the highest robustness across most scenarios. For ResNet-18, it achieves \textbf{47.07\%} robustness at 90\% sparsity, outperforming HARP and TwinRep by +1.97\%. Although slightly behind HARP in SAcc at 75\% sparsity, our method consistently performs well across sparsity levels. For VGG-16, it achieves \textbf{46.44\%} robustness at 90\% sparsity, a +2.19\% improvement over TwinRep. 

\paragraph{Key Insights} Our method excels in achieving a robust balance between adversarial robustness and accuracy, even under extreme compression. At 90\% sparsity, it consistently outperforms state-of-the-art approaches such as TwinRep and HARP. On CIFAR-10, our approach achieves notable robustness improvements of +2.14\% for ResNet-18 and +5.55\% for VGG-16, highlighting its ability to retain critical robustness under aggressive pruning. Additionally, in some cases, higher sparsity levels marginally enhance model performance, likely due to reduced overfitting. Furthermore, the method demonstrates strong generalizability across datasets and architectures, maintaining leading robustness on SVHN at high sparsity for both ResNet-18 and VGG-16. This adaptability and consistent performance underline its potential as a versatile and effective solution for robust pruning.

By effectively balancing accuracy, robustness, and sparsity, our method establishes itself as a leading approach in adversarially robust pruning, significantly outperforming existing state-of-the-art methods.

\subsection{Ablation Study}

\textbf{Impact of Adversarial Training Methods}
Our analysis demonstrates that incorporating adversarial training methods, such as TRADES, MART, and PGD, significantly enhances the robustness of pruned networks against adversarial attacks. As shown in Table~\ref{table:performance_results_updated}, TRADES consistently achieves the best balance between robustness and accuracy, while MART prioritizes SAcc with some compromise on defense. PGD provides a baseline improvement but lags behind in overall performance. Despite the notable robustness gains across all methods, adversarial training often results in a slight reduction in SAcc, highlighting the inherent trade-off between clean performance and adversarial resilience.

\begin{table}[t!]
\caption{Performance of Different Adversarial Training Methods for CIFAR-10 and SVHN.}
\label{table:performance_results_updated}
\centering
\resizebox{0.8\columnwidth}{!}{%
\begin{tabular}{l l c c c | c c c}
\toprule
\multirow{2}{*}{Arch} & \multirow{2}{*}{Method} & \multicolumn{3}{c|}{CIFAR-10} & \multicolumn{3}{c}{SVHN} \\
\cmidrule(lr){3-5} \cmidrule(lr){6-8}
 &  & FLOPs $\downarrow$ & SAcc & Adv$_{\text{AA}}$ & FLOPs $\downarrow$ & SAcc & Adv$_{\text{AA}}$ \\
\midrule

\multirow{4}{*}{ResNet} 
& Dense & 0 & 92.26 & 10.96 & 0 & 97.01 & 14.21 \\
& TRADES & 74.47 & 83.23 & 51.95 & 76.88 & 91.62 & 46.21 \\
& MART & 74.47 & 83.35 & 40.47 & 76.88 & 89.83 & 39.52 \\
& PGD & 74.47 & 85.76 & 30.45 & 76.88 & 94.44 & 34.60 \\

\midrule

\multirow{4}{*}{VGG} 
& Dense & 0 & 93.59 & 12.75 & 0 & 96.90 & 14.56 \\
& TRADES & 79.83 & 84.46 & 51.18 & 73.48 & 90.11 & 46.37 \\
& MART & 79.83 & 86.53 & 46.58 & 73.48 & 90.16 & 43.62 \\
& PGD & 79.83 & 88.12 & 37.10 & 73.48 & 94.50 & 36.30 \\

\midrule

\multirow{4}{*}{Mobilevit} 
& Dense & 0 & 92.96 & 18.51 & 0 & 96.90 & 14.56 \\
& TRADES & 73.92 & 80.49 & 43.07 & 73.39 & 88.24 & 36.35 \\
& MART & 73.92 & 80.28 & 38.17 & 73.39 & 88.49 & 34.56 \\
& PGD & 73.92 & 80.35 & 26.68 & 73.39 & 91.13 & 40.91 \\

\bottomrule
\end{tabular}%
}
\end{table}

\begin{table}[t!]
\caption{Performance of Different Pruning Criteria for CIFAR-10 and SVHN.}
\label{table:performance_diff_criteria}
\centering
\resizebox{0.8\columnwidth}{!}{%
\begin{tabular}{l l c c c | c c c}
\toprule
\multirow{2}{*}{Arch} & \multirow{2}{*}{Criterion} & \multicolumn{3}{c|}{CIFAR-10} & \multicolumn{3}{c}{SVHN} \\
\cmidrule(lr){3-5} \cmidrule(lr){6-8}
 &  & FLOPs $\downarrow$ & SAcc & Adv$_{\text{AA}}$ & FLOPs $\downarrow$ & SAcc & Adv$_{\text{AA}}$ \\
\midrule

\multirow{3}{*}{ResNet} 
& Magnitude & 52.56 & 83.58 & 48.78 & 52.21 & 91.77 & 46.92 \\
& Taylor & 52.56 & 83.23 & 48.76 & 52.21 & 91.03 & 45.77 \\
& LAMP & 52.56 & 85.17 & 47.96 & 52.21 & 91.47 & 46.15 \\

\midrule

\multirow{3}{*}{VGG} 
& Magnitude & 58.30 & 84.52 & 55.39 & 53.27 & 91.45 & 47.58 \\
& Taylor & 58.30 & 84.85 & 53.35 & 53.27 & 90.74 & 46.97 \\
& LAMP & 58.30 & 84.47 & 53.24 & 53.27 & 90.86 & 46.90 \\

\bottomrule
\end{tabular}%
}
\end{table}

\textbf{Impact of Pruning Criteria}
To evaluate the influence of different pruning criteria, we compare Magnitude, Taylor, and LAMP-based pruning strategies in Table \ref{table:performance_diff_criteria}. The results indicate that while different criteria lead to minor variations in standard and adversarial accuracy, they do not fundamentally alter the final performance trends. This suggests that the choice of pruning criterion is less critical than the overall pruning strategy and subsequent adversarial fine-tuning. Regardless of the criterion used, the effectiveness of robustness preservation is largely dictated by the adaptive pruning ratios and adversarial fine-tuning, reinforcing the key insight that structured pruning alone is insufficient to maintain robustness without a carefully designed, iterative fine-tuning process. More detailed results can be found in Appendix \ref{appendix_criteria_impact}.

\begin{table}[t!]
\caption{Performance of Different Fine-tuning Methods on ResNet with FLOPs Reduction for Tiny-ImageNet.}
\label{table: appendix_perf_resnet_tiny_imagenet}
\centering  
\setlength{\tabcolsep}{3pt}  
\resizebox{0.7\columnwidth}{!}{  
\begin{tabular}{l l c c c c}
\toprule
Architecture & Method & FLOPs $\downarrow$ & SAcc & Adv$_{\text{PGD}}$ & Adv$_{\text{FGSM}}$ \\
\midrule

\multirow{2}{*}
& Dense & 0 & 62.40 & 9.28 & 12.88 \\
\cmidrule(lr){2-6}
& Taylor & 74.83 & 57.01 & 7.46 & 10.09 \\
& Taylor + AT & 74.83 & 57.52 & 21.30 & 26.64 \\
& Taylor + MRPF & 72.62 & 59.63 & 25.26 & 28.53 \\
& Magnitude & 74.83 & 57.64 & 8.11 & 10.48 \\
& Magnitude + AT & 74.83 & 57.56 & 21.25 & 25.68 \\{ResNet-50} 
& Magnitude + MRPF & 72.62 & 58.69 & 23.94 & 26.70 \\ 

\cmidrule(lr){2-6}
& Taylor & 84.03 & 56.29 & 8.67 & 11.31 \\
& Taylor + AT & 84.03 & 55.55 & 19.32 & 24.34 \\
& Taylor + MRPF & 88.77 & 55.96 & 20.82 & 25.32 \\
& Magnitude & 84.03 & 56.59 & 6.01 & 8.44 \\
& Magnitude + AT & 84.03 & 55.76 & 19.78 & 24.86 \\
& Magnitude + MRPF & 88.77 & 55.34 & 20.47 & 25.78 \\

\bottomrule
\end{tabular}
}
\end{table}

\paragraph{Effectiveness of MRPF on Simplified Adversarial Training}  
The results in \autoref{table: appendix_perf_resnet_tiny_imagenet} highlight a key distinction: while different pruning criteria lead to minimal performance variation, adversarial training plays a decisive role in improving robustness. Even the simplest adversarial training (AT) significantly enhances adversarial accuracy, demonstrating its necessity in the pruning pipeline. MRPF further amplifies this effect, consistently yielding higher robustness compared to standard AT-based fine-tuning. These findings reinforce that pruning alone is insufficient for robustness preservation—without adversarially-aware fine-tuning, pruned models remain highly vulnerable. The effectiveness of MRPF across different adversarial training settings and datasets, including Tiny-ImageNet, underscores its adaptability in balancing efficiency and robustness.

\section{Conclusion and Future Work}

This paper presented the Module Robust Pruning and Fine-Tuning (MRPF) framework, which addresses the challenge of balancing model compression and adversarial robustness. Our work demonstrated that fine-tuning, particularly adversarial training, plays a pivotal role in restoring and enhancing both accuracy and robustness in pruned neural networks, surpassing the impact of specific pruning criteria. By introducing the Module Robustness Sensitivity (MRS) metric, MRPF adaptively adjusts layer-wise pruning ratios to prioritize robustness-critical parameters. 

Extensive experiments across diverse datasets (CIFAR-10, CIFAR-100, SVHN, and Tiny-ImageNet) and architectures (ResNet, VGG, MobileViT) validated the effectiveness of MRPF in significantly improving adversarial robustness under attacks such as PGD, FGSM, and AA, while maintaining competitive accuracy and computational efficiency. These results underscore MRPF's scalability and adaptability to different architectures and pruning settings.

Looking ahead, our method's demonstrated adaptability to transformer-based models such as MobileViT suggests its potential for application to larger-scale architectures, including Vision Transformers and large language models. This will allow us to evaluate MRPF's robustness and efficiency in handling more complex scenarios, such as high-dimensional input spaces or extremely deep networks. Furthermore, we plan to explore adaptive adversarial training strategies tailored to specific architectural properties, aiming to optimize the trade-offs between robustness, accuracy, and computational cost. These efforts will provide deeper insights into the applicability and limitations of MRPF, paving the way for more robust and efficient pruning frameworks in diverse deep learning applications.

\bibliographystyle{unsrtnat}
\bibliography{references}  





\newpage
\onecolumn
\appendix

\newpage
\appendix
\onecolumn
\section{Algorithm of MRPF} \label{appendix_algorithm}
The complete algorithm of Module Robust Pruning and Fine-Tuning (MRPF) is presented in Algorithm \ref{alg:mrpf}.

\begin{algorithm}[H]
    \caption{Module Robust Pruning and Fine-Tuning (MRPF)}
    \label{alg:mrpf}
    \begin{algorithmic}[1]
        \REQUIRE Training data \( D \); pre-trained neural network \( N \) with layers \( \{ l_i \} \), each with \( k_i \) channels; global pruning ratio \( r_g \); max per-layer pruning ratio \( r_{\text{max}} \); min per-layer pruning ratio \( r_{\text{min}} \); adversarial attack method \( AT \); perturbation limit \( \epsilon \); learning rate \( \eta \); small constant \( \delta \); batch size \( B \); number of epochs \( E \).
        \ENSURE Pruned and fine-tuned neural network \( N' \).

        \STATE \textbf{1. Generate Adversarial Examples:} Generate adversarial examples \( AE \) using \( AT \) and compute the baseline adversarial loss:
        \[
        L_{\text{orig}} = \frac{1}{|AE|} \sum_{(x, y) \in AE} \ell(N(x), y).
        \]

        \STATE \textbf{2. Calculate MRS for Each Layer:}
        \FOR{each layer \( l_i \) in \( N \)}
            \STATE Clone the network \( N_i \) and enable gradients for weights \( W_{l_i} \), freezing all other layers.
            \STATE Store the original weights \( W_{l_i}^{\text{orig}} \).
            \FOR{epoch \( e = 1 \) to \( E \)}
                \FOR{each batch \( (X_{\text{adv}}, y_{\text{adv}}) \) in \( AE \)}
                    \STATE Compute adversarial loss \( L_{\text{adv}} \) and perform gradient ascent:
                    \[
                    W_{l_i} \leftarrow W_{l_i} + \eta \nabla_{W_{l_i}} L_{\text{adv}}.
                    \]
                    \STATE Project weights back to the constrained region:
                    \[
                    \| W_{l_i} - W_{l_i}^{\text{orig}} \|_2 \leq \epsilon \| W_{l_i}^{\text{orig}} \|_2.
                    \]
                \ENDFOR
            \ENDFOR
            \STATE Compute the perturbed adversarial loss:
            \[
            L_{\text{pert}} = \frac{1}{|AE|} \sum_{(x, y) \in AE} \ell(N_i(x), y).
            \]
            \STATE Calculate the Module Robustness Sensitivity (MRS):
            \[
            \text{MRS}(l_i) = L_{\text{pert}} - L_{\text{orig}}.
            \]
        \ENDFOR

        \STATE \textbf{3. Compute Pruning Ratios:} Normalize MRS values across all layers and compute pruning ratios:
        \[
        p_i = \min \left( r_g \cdot \frac{\text{InvMRS}(l_i)}{\sum_j \text{InvMRS}(l_j)}, \, r_{\text{max}} \right),
        \]
        where \( \text{InvMRS}(l_i) = \frac{1}{\text{MRS}(l_i) + \delta} \).

        \STATE \textbf{4. Prune the Network:} 
        \FOR{each layer \( l_i \) in \( N \)}
            \STATE Rank channels by importance scores (e.g., magnitude or Taylor expansion).
            \STATE Prune the lowest \( \lfloor p_i \cdot k_i \rfloor \) channels.
        \ENDFOR
        \STATE Obtain the pruned network \( N' \).

        \STATE \textbf{5. Fine-tune the Pruned Model:}
        Generate a new set of adversarial examples \( AE' = AT(D) \). Fine-tune the pruned network \( N' \) on a dataset \( D' \) augmented with \( AE' \) using adversarial training (e.g., TRADES).

        RETURN Pruned and fine-tuned neural network \( N' \).
    \end{algorithmic}
\end{algorithm}

\section{EXPERIMENT DETAILS} \label{appendix_exp}

\subsection{Datasets and Model Architectures} \label{sec:datasets_models}

\subsubsection{Datasets.} We evaluate MRPF across four widely-used image classification datasets with varying levels of complexity:
\begin{itemize}
    \item \textbf{CIFAR-10 and CIFAR-100~\citep{Krizhevsky09}:} CIFAR-10 consists of 60,000 32x32 color images distributed across 10 classes, with distinct features for each class. CIFAR-100, a more challenging variant, contains the same number of images but spans 100 classes, introducing significant inter-class similarity.
    \item \textbf{SVHN~\citep{netzer2011reading}:} The Street View House Numbers dataset contains 99,289 real-world digit images, including 73,257 training and 26,032 testing examples. Its natural image backgrounds and digit overlaps make it distinct from CIFAR datasets.
    \item \textbf{Tiny-ImageNet~\citep{le2015tiny}:} A subset of the ImageNet dataset, Tiny-ImageNet contains 100,000 images resized to 64x64 pixels, spanning 200 classes. This dataset increases classification complexity with higher image resolution and greater inter-class similarity.
\end{itemize}

\paragraph{Model Architectures.} We evaluate MRPF on three distinct neural network architectures to cover a range of network types:
\begin{itemize}
    \item \textbf{ResNet~\citep{he2016identity}:} ResNet-18 and ResNet-50 are used as representatives of deep residual networks.
    \item \textbf{VGG-16~\citep{simonyan2014very}:} A classical convolutional architecture with a uniform layer structure and fixed filter sizes.
    \item \textbf{MobileViT-xs~\citep{mehta2021mobilevit}:} A lightweight hybrid model combining convolutional layers and vision transformers, designed for resource-efficient deployment.
\end{itemize}

These datasets and models provide a robust framework to evaluate MRPF across varying levels of complexity and model design.

\subsection{Adversarial Example Generation Methods}

\textbf{FGSM}~\citep{goodfellow2014explaining} is a single-step attack that generates adversarial examples by adding a perturbation to the input in the direction that maximizes the loss. Given a neural network model $N$ with parameters $\theta$, an input sample $\mathbf{x}$, and its true label $y$, the adversarial example $\mathbf{x}^{\text{adv}}$ is generated as:

\begin{equation}
\label{eq:fgsm}
\mathbf{x}^{\text{adv}} = \mathbf{x} + \epsilon \cdot \text{sign}\left( \nabla_{\mathbf{x}} \mathcal{L}\left( N(\mathbf{x}; \theta), y \right) \right),
\end{equation}

where:
\begin{itemize}
    \item $\epsilon$ is the perturbation magnitude, controlling the strength of the attack.
    \item $\mathcal{L}(\cdot, \cdot)$ is the loss function used for training (e.g., cross-entropy loss).
    \item $\nabla_{\mathbf{x}} \mathcal{L}\left( N(\mathbf{x}; \theta), y \right)$ is the gradient of the loss with respect to the input $\mathbf{x}$.
    \item $\text{sign}(\cdot)$ denotes the element-wise sign function.
\end{itemize}

The perturbation $\epsilon \cdot \text{sign}\left( \nabla_{\mathbf{x}} \mathcal{L} \right)$ is designed to maximize the loss, effectively causing the model to misclassify the input while keeping the perturbation minimal and often imperceptible to humans.

\textbf{PGD}~\citep{madry2017towards} is an iterative attack that refines the adversarial example over multiple steps, making it a stronger attack than FGSM. Starting from the original input $\mathbf{x}^0 = \mathbf{x}$, PGD updates the adversarial example by:

\begin{equation}
\label{eq:pgd}
\mathbf{x}^{t+1} = \Pi_{\mathcal{B}(\mathbf{x}, \epsilon)} \left( \mathbf{x}^t + \alpha \cdot \text{sign}\left( \nabla_{\mathbf{x}} \mathcal{L}\left( N(\mathbf{x}^t; \theta), y \right) \right) \right),
\end{equation}

for $t = 0, 1, \dots, T-1$, where:
\begin{itemize}
    \item $\alpha$ is the step size for each iteration.
    \item $T$ is the total number of iterations.
    \item $\Pi_{\mathcal{B}(\mathbf{x}, \epsilon)}(\cdot)$ is the projection operator that ensures $\mathbf{x}^{t+1}$ stays within the $\epsilon$-ball $\mathcal{B}(\mathbf{x}, \epsilon)$ around the original input $\mathbf{x}$ and within the valid data range (e.g., $[0, 1]$ for image data).
\end{itemize}

After $T$ iterations, the final adversarial example is $\mathbf{x}^{\text{adv}} = \mathbf{x}^T$. The PGD attack can be seen as a multi-step variant of FGSM, repeatedly applying small perturbations and projecting back into the feasible set to find a more effective adversarial example.

\textbf{AutoAttack(AA)}~\citep{croce2020reliable} is a composite adversarial attack framework that combines multiple methods, including an enhanced version of Projected Gradient Descent (PGD) known as Adaptive PGD (APGD). APGD dynamically adjusts the step size and incorporates a momentum term to improve attack effectiveness. The adversarial example $\mathbf{x}^{\text{adv}}$ in APGD is generated iteratively as follows:

\begin{equation}
\label{eq:apgd_update}
\begin{aligned}
    z^{(k+1)} &= \Pi_S\left( \mathbf{x}^{(k)} + \eta^{(k)} \nabla_{\mathbf{x}} \mathcal{L}\left( N(\mathbf{x}^{(k)}; \theta), y \right) \right), \\
    \mathbf{x}^{(k+1)} &= \Pi_S\left( \mathbf{x}^{(k)} + \alpha \cdot \left( z^{(k+1)} - \mathbf{x}^{(k)} \right) + (1 - \alpha) \cdot \left( \mathbf{x}^{(k)} - \mathbf{x}^{(k-1)} \right) \right),
\end{aligned}
\end{equation}

where:
\begin{itemize}
    \item $\Pi_S(\cdot)$ is the projection operator that ensures the adversarial example stays within the feasible set $S$ (e.g., $\epsilon$-ball and valid data range).
    \item $\eta^{(k)}$ is the step size at iteration $k$, which is adaptively adjusted during optimization.
    \item $\alpha$ is a momentum term (typically set to $\alpha = 0.75$) that balances the influence of previous updates.
    \item $\nabla_{\mathbf{x}} \mathcal{L}\left( N(\mathbf{x}; \theta), y \right)$ is the gradient of the loss $\mathcal{L}$ with respect to the input $\mathbf{x}$.
\end{itemize}

The algorithm restarts from the best point $\mathbf{x}_{\max}$ whenever the step size is reduced, ensuring a localized search around promising regions. This restart mechanism enhances the optimization by focusing on high-loss areas.

By combining APGD with other attacks, AA provides a reliable and comprehensive evaluation of model robustness. Its adaptive mechanisms, such as step size adjustment and restarts, make it a strong and consistent adversarial attack strategy.

\subsubsection*{Implementation Details}

In our experiments, as described in Appendix~\ref{sec:hyperparams_training}, we use FGSM with $\epsilon = \frac{8}{255}$ to generate adversarial examples for both CIFAR-10 and CIFAR-100 datasets. For Tiny-ImageNet, we use the same $\epsilon$ value to maintain consistency across experiments.

For FGSM, since it is a single-step method, we set the step size $\alpha$ equal to $\epsilon$. For PGD, which we use in ablation studies to test the robustness of our pruned models, we set the step size $\alpha = \frac{\epsilon}{T}$, where $T$ is the number of iterations (we set 20 in the experiments). This ensures that the total perturbation remains within the $\epsilon$ constraint.

Integrating adversarial examples generated by FGSM and PGD into our pruning process serves multiple purposes:
\begin{itemize}
    \item \textbf{Evaluating Robustness}: By assessing the model's performance on adversarial examples, we establish a baseline for robustness, ensuring that pruning does not degrade this critical property.
    \item \textbf{Guiding Pruning Decisions}: The gradients computed during adversarial example generation highlight the model's sensitive features. By focusing on these gradients, we can identify which neurons or channels are crucial for robustness and should be retained.
    \item \textbf{Adversarial Training}: Incorporating adversarial examples into the fine-tuning process helps the pruned model recover and maintain robustness against attacks, as discussed in Section~\ref{sec:pruning_settings}.
\end{itemize}

\subsection{Adversarial Training Methods}

Adversarial training is a widely-used defense strategy designed to improve the robustness of neural networks against adversarial attacks by incorporating adversarial examples into the training process. Below, we describe three representative frameworks: PGD-based adversarial training, TRADES, and MART.

\paragraph{PGD-based Adversarial Training.}
PGD-based adversarial training~\citep{madry2017towards} constructs adversarial examples iteratively using the Projected Gradient Descent (PGD) attack and optimizes the model parameters to minimize the worst-case loss over these examples. Given an input $\mathbf{x}$, its true label $y$, and a neural network model $N$ parameterized by $\theta$, adversarial examples $\mathbf{x}^{\text{adv}}$ are generated iteratively:

\begin{equation}
\label{eq:pgd_adv}
\mathbf{x}^{(t+1)} = \Pi_{\mathcal{B}(\mathbf{x}, \epsilon)} \left( \mathbf{x}^{(t)} + \alpha \cdot \text{sign}\left( \nabla_{\mathbf{x}} \mathcal{L}\left( N(\mathbf{x}^{(t)}; \theta), y \right) \right) \right),
\end{equation}

where:
\begin{itemize}
    \item $\alpha$ is the step size.
    \item $\Pi_{\mathcal{B}(\mathbf{x}, \epsilon)}(\cdot)$ is the projection operator that ensures $\mathbf{x}^{(t+1)}$ remains within the $\epsilon$-ball $\mathcal{B}(\mathbf{x}, \epsilon)$ around $\mathbf{x}$ and the valid input range (e.g., $[0, 1]$ for image data).
\end{itemize}

The model is trained to minimize the worst-case loss under adversarial perturbations:

\begin{equation}
\label{eq:pgd_loss}
\min_{\theta} \mathbb{E}_{(\mathbf{x}, y) \sim \mathcal{D}} \left[ \max_{\mathbf{x}^{\prime} \in \mathcal{B}(\mathbf{x}, \epsilon)} \mathcal{L}(N(\mathbf{x}^{\prime}; \theta), y) \right],
\end{equation}

where $\mathcal{D}$ denotes the data distribution.

\paragraph{TRADES.}
TRADES (TRadeoff-inspired Adversarial Defense via Surrogate-loss)~\citep{zhang2019theoretically} explicitly balances the trade-off between natural accuracy and adversarial robustness by introducing a regularization term. The objective function is given as:

\begin{equation}
\label{eq:trades_loss}
\mathcal{L}_{\text{TRADES}} = \mathbb{E}_{(\mathbf{x}, y) \sim \mathcal{D}} \left[ \mathcal{L}(N(\mathbf{x}; \theta), y) + \beta \cdot \max_{\mathbf{x}^{\prime} \in \mathcal{B}(\mathbf{x}, \epsilon)} \mathcal{D}_{\text{KL}}(N(\mathbf{x}; \theta) \| N(\mathbf{x}^{\prime}; \theta)) \right],
\end{equation}

where:
\begin{itemize}
    \item $\mathcal{L}(\cdot, \cdot)$ is the classification loss (e.g., cross-entropy loss).
    \item $\mathcal{D}_{\text{KL}}(\cdot \| \cdot)$ is the Kullback-Leibler (KL) divergence between the output distributions of clean and adversarial examples.
    \item $\beta$ is a hyperparameter controlling the trade-off between robustness and accuracy.
\end{itemize}

\paragraph{MART.}
MART (Misclassification Aware adversarial Robustness Training)~\citep{wang2019improving} focuses on correctly classified examples while penalizing misclassified ones, aiming to improve both accuracy and robustness. The loss function is defined as:

\begin{equation}
\label{eq:mart_loss}
\mathcal{L}_{\text{MART}} = \mathbb{E}_{(\mathbf{x}, y) \sim \mathcal{D}} \left[ \mathcal{L}(N(\mathbf{x}^{\text{adv}}; \theta), y) + \lambda \cdot \mathbb{I}\{N(\mathbf{x}; \theta) \neq y\} \cdot \mathcal{D}_{\text{KL}}(N(\mathbf{x}; \theta) \| N(\mathbf{x}^{\text{adv}}; \theta)) \right],
\end{equation}

where:
\begin{itemize}
    \item $\mathbb{I}\{\cdot\}$ is the indicator function that activates the KL divergence term for misclassified examples.
    \item $\lambda$ is a weighting factor for the misclassification-aware penalty.
\end{itemize}

These adversarial training methods address different aspects of robustness improvement: PGD-based training focuses on worst-case loss minimization, TRADES highlights the robustness-accuracy trade-off, and MART emphasizes handling misclassified examples. Together, they form a strong foundation for designing robust machine learning models.

\subsection{Hyperparameter Settings and Training Details} \label{sec:hyperparams_training}

\subsubsection{Adversarial Training.} MRPF employs adversarial training to fine-tune pruned networks. TRADES~\citep{zhang2019theoretically} is used as the default framework, balancing SAcc and adversarial robustness. For comparative purposes, MART~\citep{wang2019improving} and PGD-based adversarial training~\citep{madry2017towards} are also evaluated. Adversarial examples are generated using PGD-10 with a perturbation bound \( \epsilon = 8/255 \) and step size \( 2/255 \). AutoAttack~\citep{croce2020reliable}, a robust benchmarking tool, is used to evaluate adversarial robustness.

\subsubsection{Pruning Strategy.}  \label{sec:pruning_settings}
Neuron importance is primarily computed using Magnitude pruning, with Taylor expansion pruning~\citep{molchanov2019importance} included for comparison. Layer-wise pruning ratios are dynamically adjusted using MRS values, normalized across the network with a global pruning ratio \( r_g \). The maximum pruning ratio per layer is capped at 80\%.

\subsubsection{Fine-tuning Details.} Fine-tuning ensures the recovery of both SAcc and adversarial robustness post-pruning. Key settings include:
\begin{itemize}
    \item \textbf{Batchsize:} 256
    \item \textbf{Optimizer:} SGD with momentum (\( \text{momentum} = 0.9 \)).
    \item \textbf{Learning Rate:} Cosine annealing schedule, starting at \( 0.02 \) for CIFAR and SVHN datasets and \( 0.1 \) for Tiny-ImageNet, decaying to \( 1e^{-4} \) over 90 epochs.This is in contrast to the pre-trained model, which was trained for 500 epochs to ensure a robust initialization prior to pruning.
    
\end{itemize}

\subsubsection{Dataset-Specific Adjustments.}
\begin{itemize}
    \item \textbf{Tiny-ImageNet:} The batch size is reduced to 128 due to higher memory demands. The initial learning rate is set to 0.1 and follows the same cosine annealing decay schedule.
\end{itemize}

These hyperparameter settings and training configurations ensure the robustness and generalizability of MRPF across datasets and architectures.

\subsection{Importance Score Calculation}

In our pruning process, we compute the importance of each channel using mainly two criteria: Magnitude importance and Taylor importance. These methods help identify which channels contribute most to the model's performance and should be retained during pruning.

\subsubsection{Magnitude Importance}

Magnitude importance measures the significance of each channel based on the magnitude of its weights. For a given convolutional layer \( l_i \) with weights \( W_{l_i} \in \mathbb{R}^{N_{i+1} \times N_i \times K \times K} \) (where \( N_i \) and \( N_{i+1} \) are the numbers of input and output channels, and \( K \) is the kernel size), the importance score \( I_{l_i,j} \) for output channel \( j \) is calculated as:

\begin{equation}
\label{eq:magnitude_importance}
I_{l_i,j} = \sum_{k=1}^{N_i} \sum_{u=1}^{K} \sum_{v=1}^{K} \left| W_{l_i,j,k,u,v} \right|^p
\end{equation}

where:
\begin{itemize}
    \item \( W_{l_i,j,k,u,v} \) is the weight connecting the \( k^{\text{th}} \) input channel to the \( j^{\text{th}} \) output channel at position \( (u, v) \) in layer \( l_i \).
    \item \( p \) is the norm degree, typically set to 2 (i.e., L2-norm).
\end{itemize}

This method sums the \( p \)-th power of the absolute values of all weights in output channel \( j \), providing a scalar importance score. Channels with higher magnitude importance are considered more significant and are less likely to be pruned.

\subsubsection{Taylor Importance}

Taylor importance estimates the contribution of each channel to the loss function using a first-order Taylor expansion~\citep{molchanov2019importance}. For output channel \( j \) in layer \( l_i \), the importance score \( I_{l_i,j} \) is calculated as:

\begin{equation}
\label{eq:taylor_importance}
I_{l_i,j} = \sum_{k=1}^{N_i} \sum_{u=1}^{K} \sum_{v=1}^{K} \left| W_{l_i,j,k,u,v} \cdot \frac{\partial \ell}{\partial W_{l_i,j,k,u,v}} \right|
\end{equation}

where:
\begin{itemize}
    \item \( \ell \) is the loss function (e.g., cross-entropy loss).
    \item \( \frac{\partial \ell}{\partial W_{l_i,j,k,u,v}} \) is the gradient of the loss with respect to the weight \( W_{l_i,j,k,u,v} \).
\end{itemize}

By considering both the weights and their gradients, Taylor importance captures how sensitive the loss is to changes in each weight. This method effectively identifies channels that, when pruned, would have a significant impact on the loss, allowing us to preserve critical channels during pruning.

\subsubsection{Normalization and Pruning Decision}

After computing the importance scores using either method, we normalize the scores to ensure fair comparison across layers. The normalization can be performed using methods such as:

\begin{equation}
\label{eq:normalization}
\tilde{I}_{l_i,j} = \frac{I_{l_i,j}}{\text{mean}(\mathbf{I}_{l_i})}
\end{equation}

where \( \mathbf{I}_{l_i} \) is the vector of importance scores for all channels in layer \( l_i \). This normalization helps in balancing the pruning process across different layers with varying importance score distributions.

Channels with lower normalized importance scores \( \tilde{I}_{l_i,j} \) are considered less critical and are pruned first. By applying these importance criteria, we effectively reduce the model size while maintaining performance and robustness.

\subsection{Module Robustness Sensitivity Calculation}
\label{appendix:mrs_calculation}

In this appendix, we provide additional details on the calculation of Module Robustness Sensitivity (MRS) used in our pruning algorithm, as outlined in Section 3. The MRS quantifies the impact of perturbing each layer's weights on the adversarial loss, guiding the pruning process to preserve layers that are crucial for robustness.

\subsubsection{Implementation Details}

To compute the MRS for each layer, we perform the following steps:

\begin{enumerate}
    \item \textbf{Perturbing Layer Weights:} For each prunable layer \( l_i \), we perform gradient ascent on the adversarial loss to perturb its weights \( W_{l_i} \), while keeping the weights of other layers fixed. The perturbation is constrained to a small norm \( \epsilon \) relative to the original weights, as described in Equation~(4) in the main text.

    \item \textbf{Computing Perturbed Adversarial Loss:} After perturbing \( W_{l_i} \), we evaluate the adversarial loss \( L_{\text{adv}} \) on the adversarial examples \( AE \) using the network with perturbed weights in layer \( l_i \).

    \item \textbf{Calculating MRS:} The MRS for layer \( l_i \) is calculated as the difference between the perturbed adversarial loss and the original adversarial loss \( L_{\text{orig}} \), as defined in Equation~(5) in the main text.

    \item \textbf{Adjustment of MRS Values:} To handle any non-positive MRS values, we replace any \( \text{MRS}(l_i) \leq 0 \) with a small positive constant \( \delta \) to prevent numerical issues during pruning ratio computation.

\end{enumerate}

\subsubsection{Pruning Ratio Computation}

Following the calculation of MRS for all layers, we compute the pruning ratios \( p_i \) for each layer. Instead of directly using the inverse of MRS values as in Equation~(6) in the main text, we normalize the MRS deviations to ensure the pruning ratios are within the desired range. The steps are as follows:

\begin{enumerate}
    \item Compute the mean MRS: \( \mu_{\text{MRS}} = \frac{1}{L} \sum_{i=1}^{L} \text{MRS}(l_i) \).

    \item Compute the deviation of MRS values from the mean: \( \Delta_i = \text{MRS}(l_i) - \mu_{\text{MRS}} \).

    \item Normalize the deviations to the range \([-1, 1]\): \( \Delta_i = \frac{\Delta_i}{\max_j |\Delta_j|} \).

    \item Compute the initial pruning ratios: \( p_i = r_g - \Delta_i \times (r_{\text{max}} - r_{\text{min}}) \).

    \item Clip the pruning ratios to be within \([r_{\text{min}}, r_{\text{max}}]\).

    \item Adjust the pruning ratios to match the target global pruning ratio \( r_g \): \( p_i \leftarrow p_i \times \frac{r_g}{r_{\text{actual}}} \), where \( r_{\text{actual}} = \frac{1}{L} \sum_{i=1}^{L} p_i \).

\end{enumerate}

This method ensures that layers with higher MRS values (more important for robustness) receive lower pruning ratios, while layers with lower MRS values are pruned more aggressively.

\subsubsection{Additional Notes}

\begin{itemize}
    \item The perturbation limit \( \epsilon \) is set to a small value (e.g., \( \epsilon = 8/255 \)) to ensure that the weight perturbations remain within a reasonable range.

    \item The learning rate \( \eta \) and the number of epochs \( E \) for the gradient ascent are chosen to effectively perturb the weights without excessive computation.

    \item The small constant \( \delta \) used to adjust non-positive MRS values is set to a small positive value (e.g., \( \delta = 1 \times 10^{-6} \)).

\end{itemize}

\section{ADDITIONAL RESULTS}\label{appendix_mrpf_res}
\subsection{MRS Scores Distribution Across Layers}
\begin{figure}[t!]
    \centering
    \includegraphics[width=0.9\linewidth]{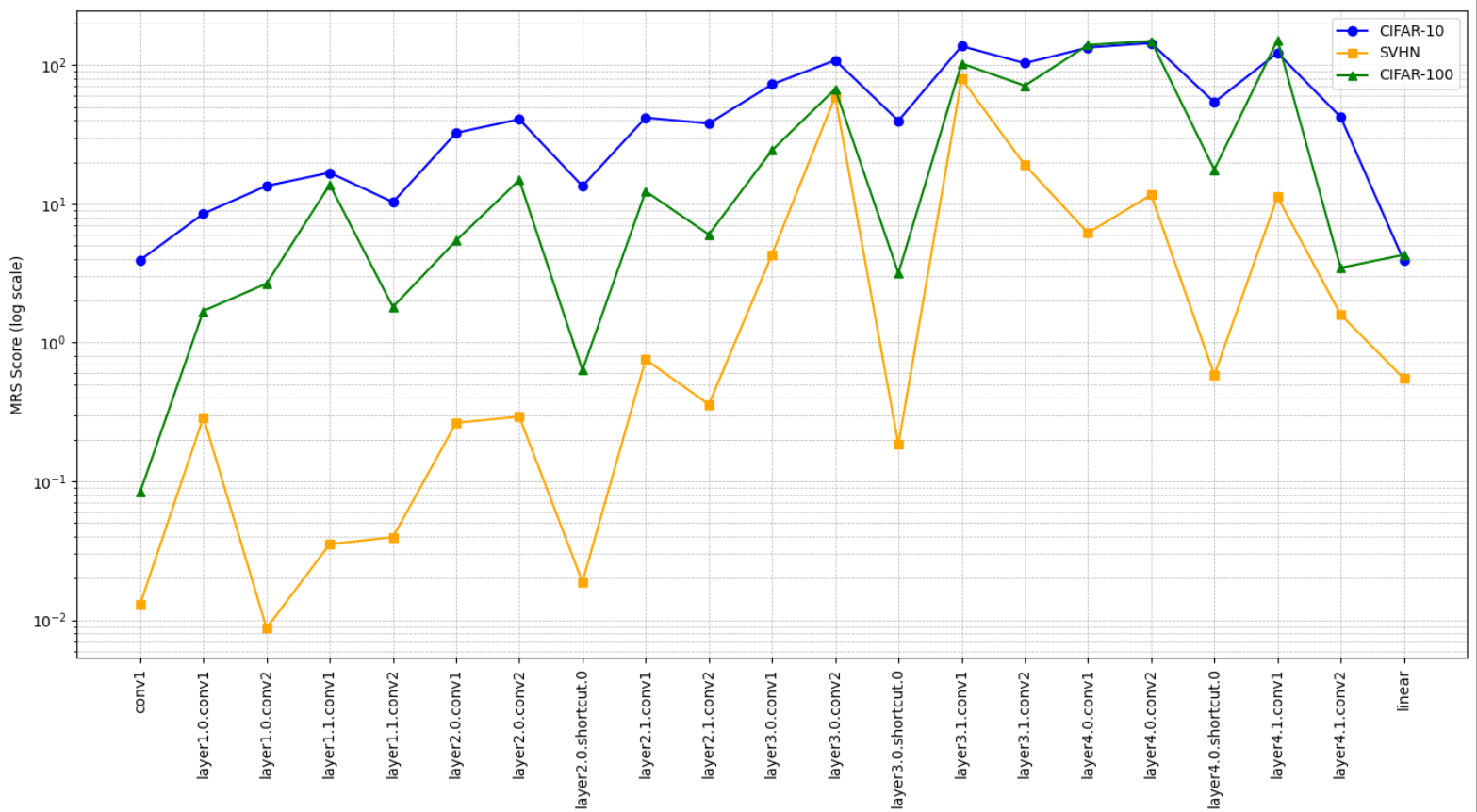} 
    \captionsetup{font=small} 
    \caption{
    MRS (Module Robustness Sensitivity) scores distribution across layers of ResNet-18 evaluated on CIFAR-10, CIFAR-100, and SVHN datasets. The y-axis uses a logarithmic scale to emphasize the variations in MRS values across layers, revealing the consistent trend of robustness-critical layers across datasets. Deeper layers generally exhibit higher sensitivity, underscoring their significant role in maintaining adversarial robustness, while early and intermediate layers display lower but crucial sensitivity to adversarial perturbations.
}
    \label{fig:MRS_resnet18}
\end{figure}
To analyze the robustness sensitivity of different layers in neural networks, we evaluated Module Robustness Sensitivity (MRS) scores for ResNet-18 across CIFAR-10, CIFAR-100, and SVHN datasets, as shown in Figure~\ref{fig:MRS_resnet18}. Despite variations in specific MRS values due to dataset differences, the overall trend of robustness-critical layers remains consistent across datasets. Deeper layers generally exhibit higher MRS values, indicating greater sensitivity to adversarial perturbations and stronger contributions to robustness, while early and intermediate layers show lower sensitivity, yet retain essential roles.

The logarithmic scale highlights subtle yet important differences, especially in early and intermediate layers, which, though less sensitive, still play a critical role in overall robustness by contributing to feature extraction and stability. These findings suggest that the importance of specific layers is largely architecture-driven, not dataset-specific, supporting the necessity of adaptive, layer-specific pruning strategies. Such strategies should prioritize robustness-critical layers, particularly deeper ones, while providing flexibility in pruning less sensitive layers to optimize the trade-off between robustness, efficiency, and compression in compact models.

\subsection{Pruning Criteria Impact} \label{appendix_criteria_impact}
\begin{figure}[t]
    \centering
    \begin{subfigure}{0.31\linewidth}
        \centering 
        \includegraphics[width=\linewidth]{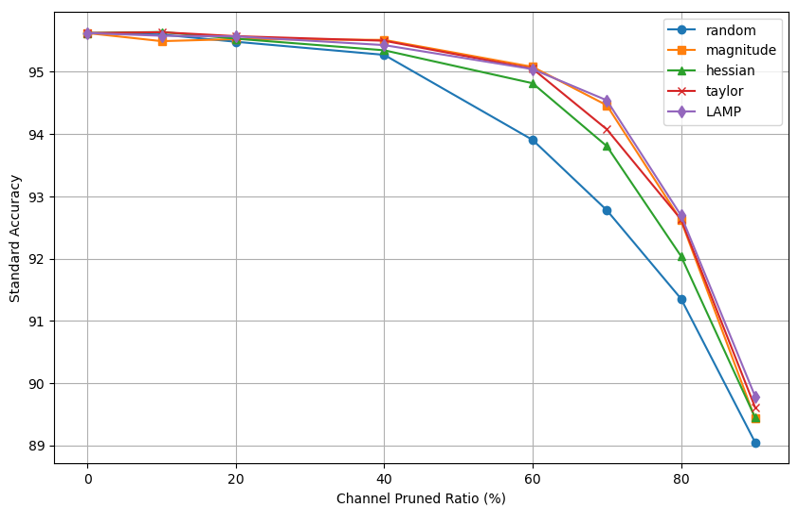}
        \captionsetup{font=tiny} 
        \caption{ResNet-50 on CIFAR-10 (Acc)}
        \label{fig:resnet50_cifar10_acc}
    \end{subfigure}
    \hfill
    \begin{subfigure}{0.32\linewidth}
        \centering
        \includegraphics[width=\linewidth]{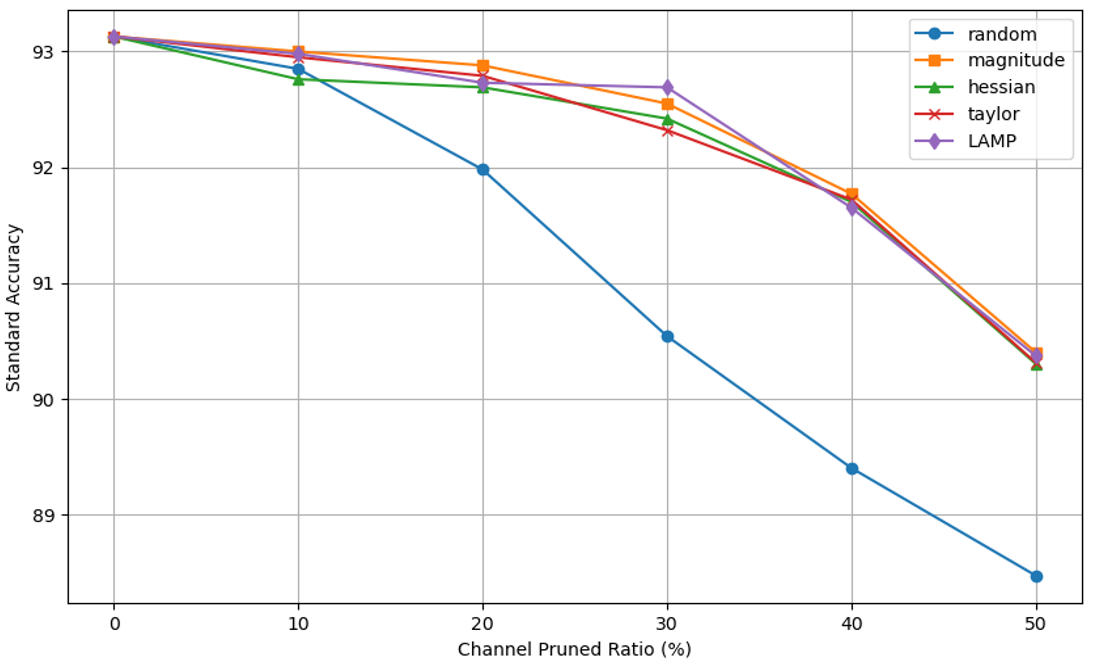}
        \captionsetup{font=tiny} 
        \caption{MobileVit-xs on CIFAR-10 (Acc)}
        \label{fig:mobilevitxs_cifar10_acc}
    \end{subfigure}
    \hfill
    \begin{subfigure}{0.33\linewidth}
        \centering
        \includegraphics[width=\linewidth]{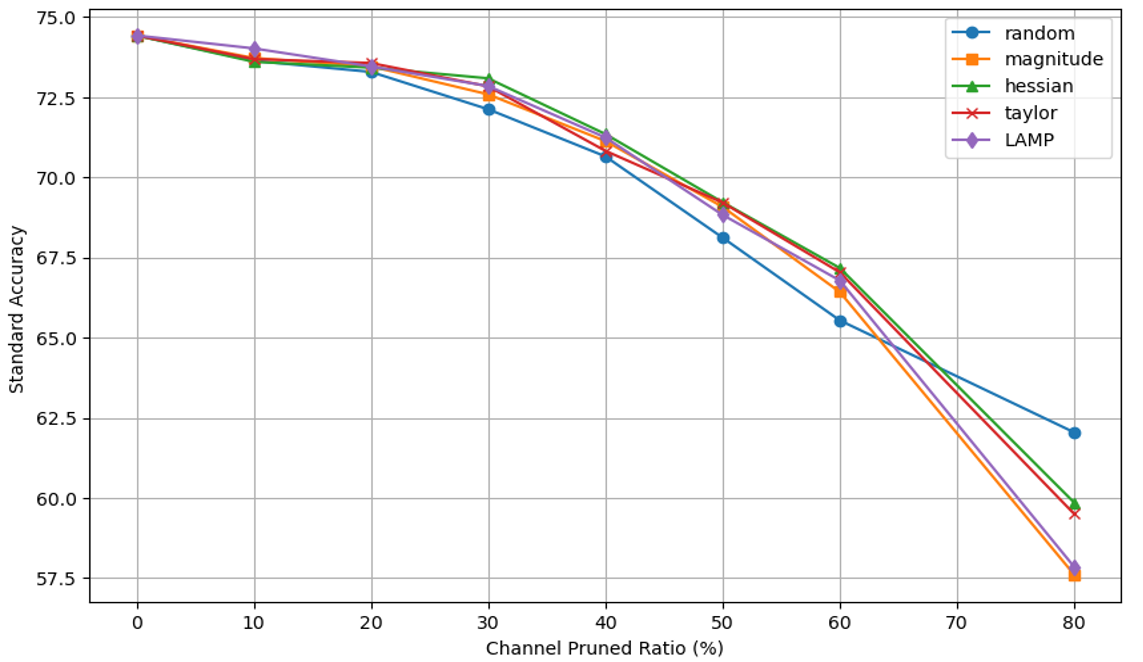}
        \captionsetup{font=tiny} 
        \caption{VGG-16 on CIFAR-100 (Acc)}
        \label{fig:vgg16_cifar100_acc}
    \end{subfigure}

    \vspace{10pt}

    \begin{subfigure}{0.32\linewidth}
        \centering
        \includegraphics[width=\linewidth]{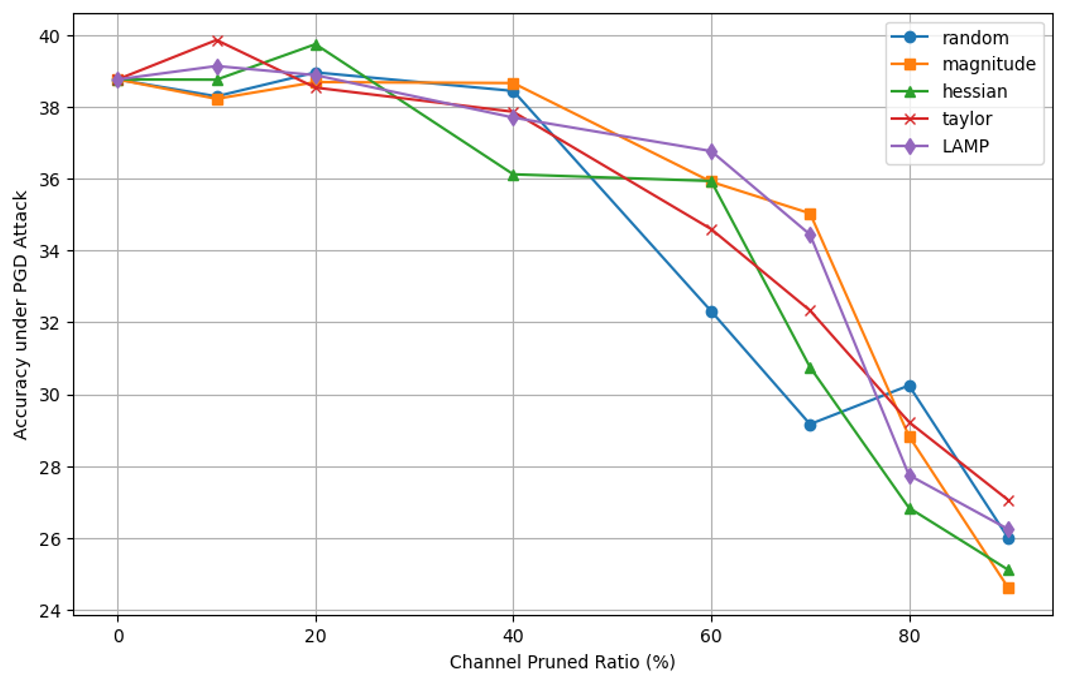}
        \captionsetup{font=tiny} 
        \caption{ResNet-50 on CIFAR-10 (PGD Attack)}
        \label{fig:resnet50_cifar10_pgd}
    \end{subfigure}
    \hfill
    \begin{subfigure}{0.33\linewidth}
        \centering
        \includegraphics[width=\linewidth]{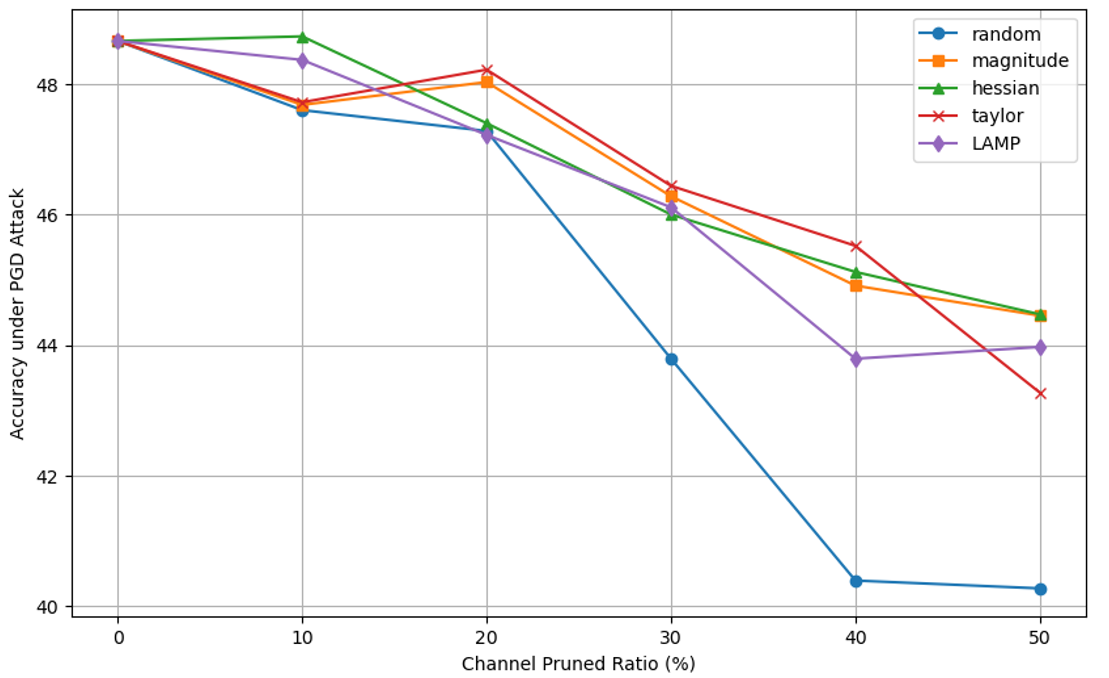}
        \captionsetup{font=tiny} 
        \caption{MobileVit-xs on CIFAR-10 (PGD Attack)}
        \label{fig:mobilevitxs_cifar10_pgd}
    \end{subfigure}
    \hfill
    \begin{subfigure}{0.33\linewidth}
        \centering
        \includegraphics[width=\linewidth]{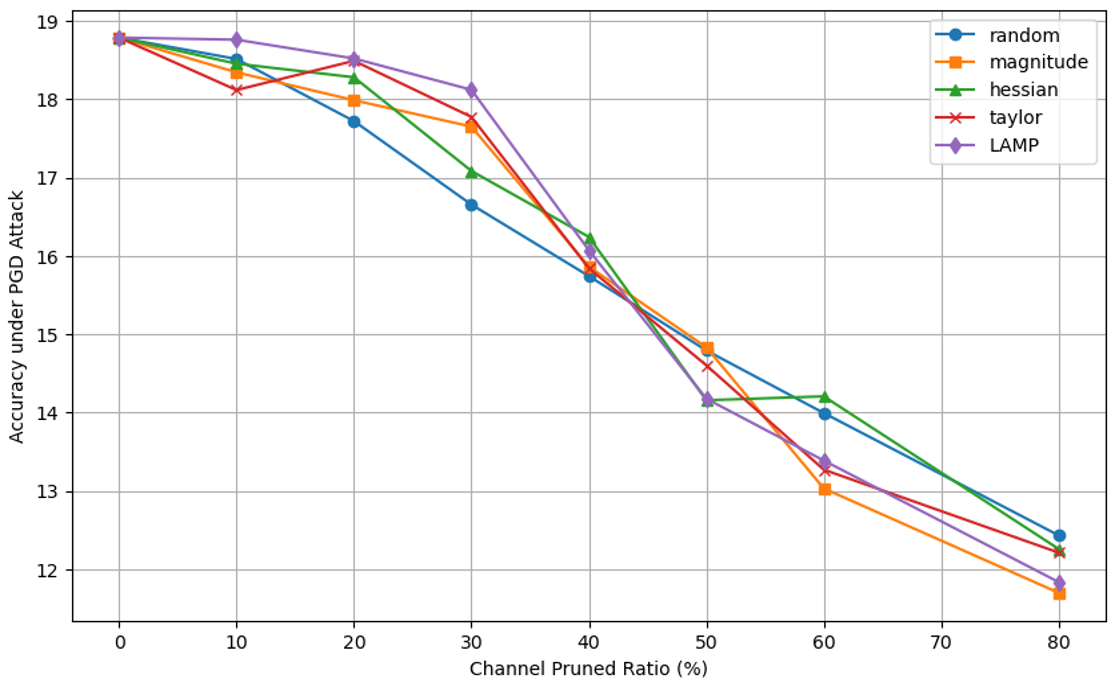}
        \captionsetup{font=tiny} 
        \caption{VGG-16 on CIFAR-100 (PGD Attack)}
        \label{fig:vgg16_cifar100_pgd}
    \end{subfigure}

    \vspace{10pt}

    \begin{subfigure}{0.32\linewidth}
        \centering
        \includegraphics[width=\linewidth]{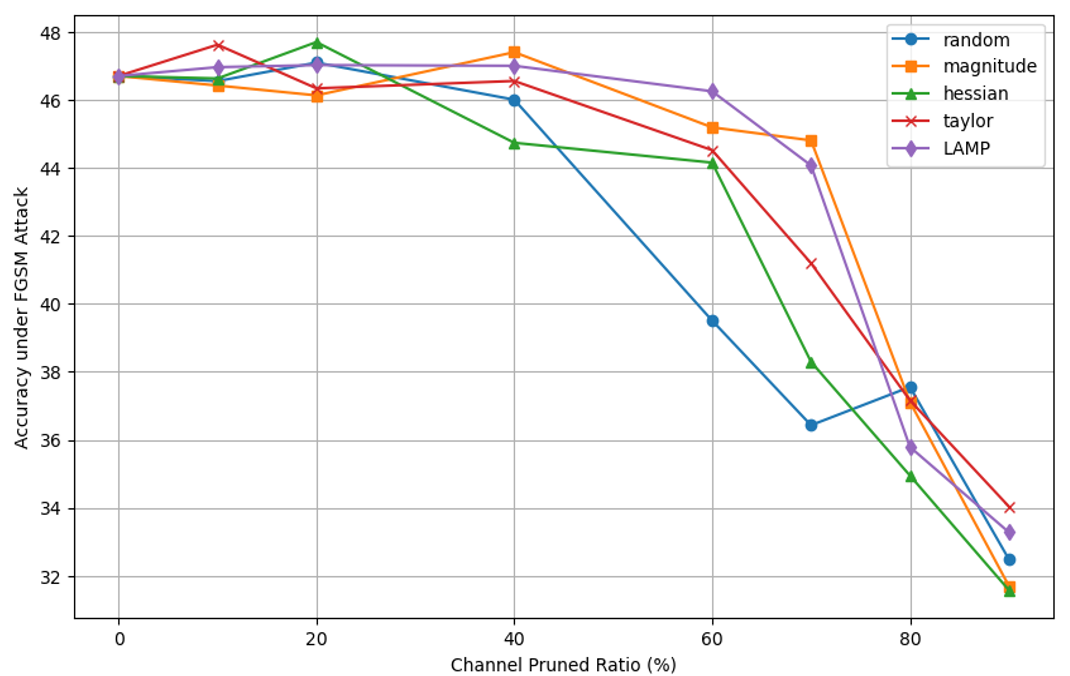}
        \captionsetup{font=tiny} 
        \caption{ResNet-50 on CIFAR-10 (FGSM Attack)}
        \label{fig:resnet50_cifar10_fgsm}
    \end{subfigure}
    \hfill
    \begin{subfigure}{0.33\linewidth}
        \centering
        \includegraphics[width=\linewidth]{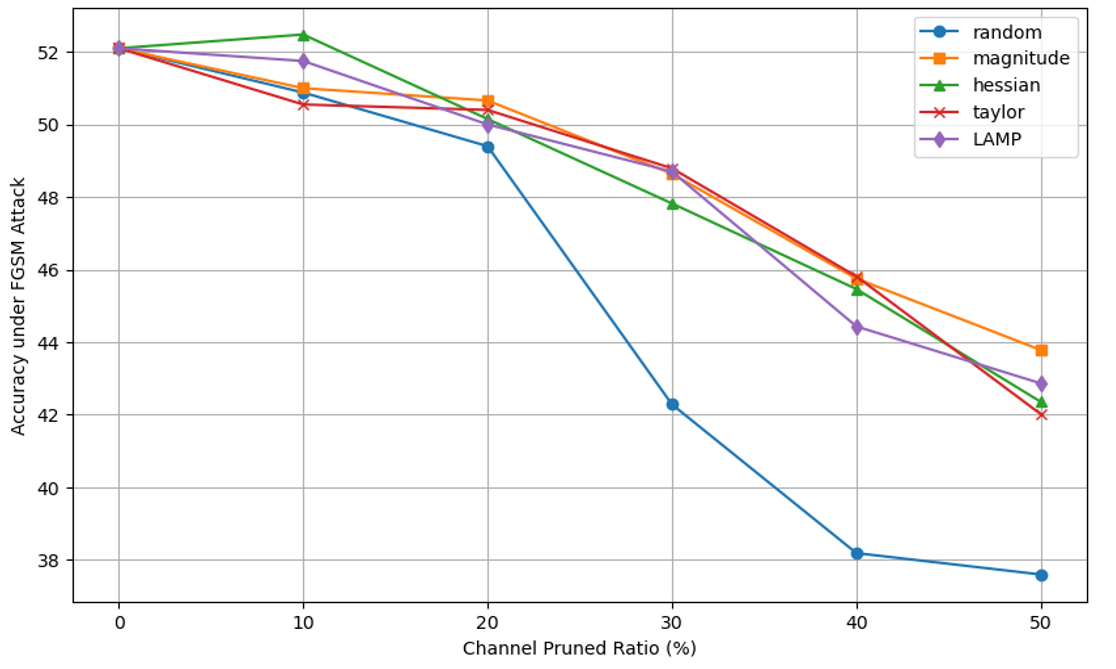}
        \captionsetup{font=tiny} 
        \caption{MobileVit-xs on CIFAR-10 (FGSM Attack)}
        \label{fig:mobilevitxs_cifar10_fgsm}
    \end{subfigure}
    \hfill
    \begin{subfigure}{0.335\linewidth}
        \centering
        \includegraphics[width=\linewidth]{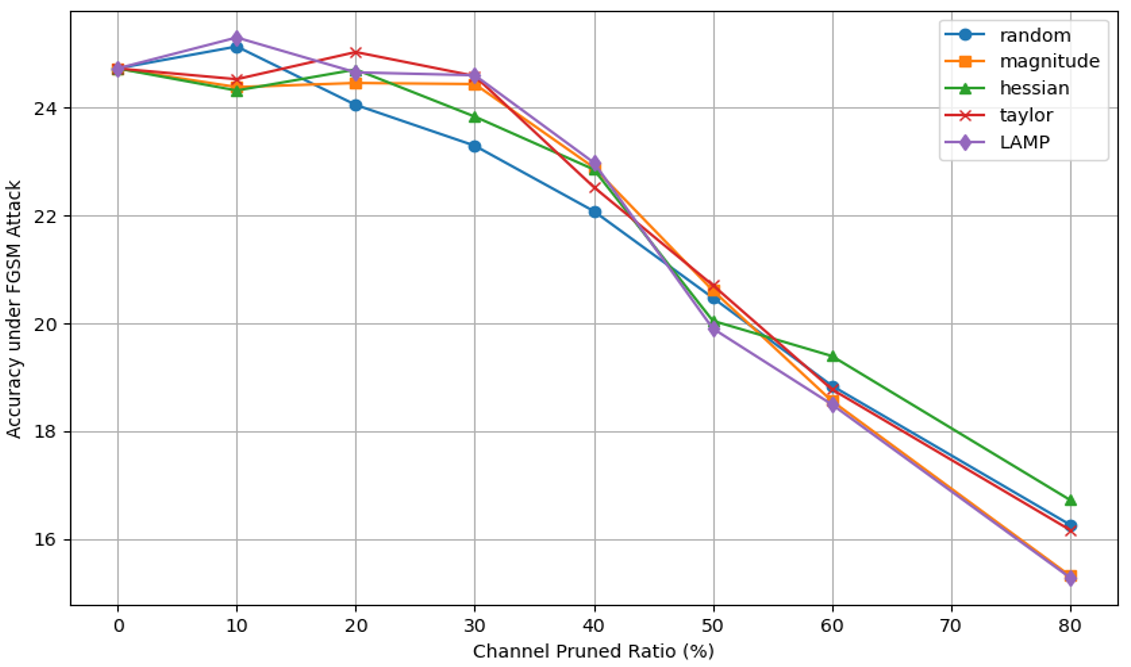}
        \captionsetup{font=tiny} 
        \caption{VGG-16 on CIFAR-100 (FGSM Attack)}
        \label{fig:vgg16_cifar100_fgsm}
    \end{subfigure}

    \caption{
        Impact of various pruning criteria on accuracy and robustness across different architectures and datasets. The figures illustrate the performance under different pruning methods (random, magnitude, Hessian, Taylor, LAMP) for ResNet-50, MobileViT-xs, and VGG-16 on CIFAR-10 and CIFAR-100. Regardless of the pruning method, similar accuracy and robustness are achieved after fine-tuning.
    }
    \label{fig:pruning_impact}
\end{figure}

 The experimental results, presented in \autoref{fig:pruning_impact}, provide several key insights into the relationship between pruning criteria, compression ratios, and the fine-tuning process. Specifically, the choice of pruning criteria, whether based on simple methods such as weight magnitude or more advanced approaches such as Hessian or Taylor series approximations, was found to have a limited impact on the final performance of pruned models after fine-tuning. This conclusion held true across all evaluated architectures, including ResNet-50, MobileViT-xs, and VGG-16, as well as across datasets like CIFAR-10 and CIFAR-100. Despite their structural differences, these models exhibited similar recovery in both standard accuracy and adversarial robustness after pruning, irrespective of the specific pruning strategy employed.

 \textbf{Minor Performance Gains at Low Compression Ratios:}
At low compression ratios (below 50\%), pruning often led to slight performance improvements. This can be attributed to the removal of redundant parameters that may have introduced noise, thus making the model more streamlined and efficient. By eliminating these unnecessary elements, the model can potentially reduce overfitting and improve accuracy. These findings suggest that moderate pruning not only reduces model size but can also enhance performance by simplifying the network.

\textbf{Performance Decline at High Compression Ratios:}
As the compression ratio increased beyond 60–70\%, both accuracy and robustness began to deteriorate across all pruning methods. This decline suggests that the overall compression level, rather than the specific pruning criteria, drives performance loss. At high compression ratios, important parameters are removed, leading to a less accurate model with coarser decision boundaries. This makes the pruned model more prone to errors and more vulnerable to adversarial attacks.

\textbf{Role of Fine-Tuning in Mitigating Pruning Effects:}
Fine-tuning was essential in recovering performance, particularly for compression levels up to 50\%. Even after pruning, models retained sufficient capacity to re-learn critical patterns and decision boundaries during retraining, restoring both accuracy and robustness. The choice of fine-tuning methods, particularly optimization algorithms and learning rate schedules, played a pivotal role in this recovery. Models pruned with different criteria but fine-tuned in a similar way typically converged to comparable performance, highlighting the dominant influence of fine-tuning.

\textbf{Limitations at Higher Compression Ratios:}
Despite the benefits of fine-tuning, its effectiveness wanes as compression ratios exceed 70\%. Even the best fine-tuning strategies cannot fully compensate for the loss of key parameters, leading to noticeable declines in both SAcc and adversarial robustness. This limitation underscores the inherent trade-off between efficiency and performance: while aggressive pruning reduces model size, it can also compromise the model’s ability to generalize and defend against adversarial perturbations.

In conclusion, while weight importance scoring is often considered pivotal in pruning strategies, the results suggest that effective fine-tuning after pruning is the dominant factor influencing the model's final performance. This observation underscores the need for a balanced approach to pruning and fine-tuning to maintain both accuracy and robustness, particularly at higher compression levels.

\subsection{Additional ablation Study}

\begin{table}[t!]
\caption{Impact of MRPF on Robustness.}
\label{table: MRPF impact}
\centering  
\setlength{\tabcolsep}{4pt} 
\resizebox{0.65\columnwidth}{!}{ 
\begin{tabular}{l l c c c c}
\toprule
Architecture & Method & FLOPs $\downarrow$ & SAcc & Adv$_{\text{PGD}}$ & Adv$_{\text{FGSM}}$ \\
\midrule

\multirow{2}{*}{ResNet-18} 
& Magnitude + AT & 51.60 & 93.8 & 10.78 & 66.79 \\
& Magnitude + MRPF & 55.63 & 93.97 & 21.06 & 67.21 \\
\cmidrule(lr){2-6}

& Magnitude + AT & 91.05 & 91.68 & 13.91 & 62.67 \\
& Magnitude + MRPF & 90.73 & 91.63 & 15.8 & 62.79 \\

\midrule

\multirow{2}{*}{VGG-16}
& Magnitude + AT & 51.28 & 92.21 & 20.28 & 64.04 \\
& Magnitude + MRPF & 54.78 & 92.22 & 21.32 & 64.48 \\
\cmidrule(lr){2-6}
& Magnitude + AT & 64.14 & 91.13 & 18.86 & 62.66 \\
& Magnitude + MRPF & 68.67 & 91.33 & 19.43 & 63.13 \\

\bottomrule
\end{tabular}
}
\end{table}

\textbf{Analysis of Pruning with MRS:} \autoref{table: MRPF impact} highlights the benefits of incorporating the Module Robustness Sensitivity (MRS) into the pruning process. When comparing MRPF to the baseline approach, which applies adversarial training (AT) without MRS, we observe a clear improvement in adversarial robustness. Across both ResNet-18 and VGG-16 architectures, MRPF consistently enhances resilience against adversarial attacks, demonstrating its ability to preserve critical parameters essential for maintaining robust decision boundaries.

In addition to improving robustness, MRPF achieves these gains with only minimal changes in SAcc and computational efficiency. This indicates that the integration of MRS into the pruning process results in a more balanced trade-off between robustness and efficiency, something that adversarial training alone cannot accomplish.

\begin{table}[t!]
\centering
\begin{minipage}{0.32\textwidth} 
    \centering
    \caption{Comparison of Different Adversarial Examples on ResNet-18.}
    \label{table:adversarial_examples_comparison}
    \setlength{\tabcolsep}{2pt} 
    \resizebox{\textwidth}{!}{
    \begin{tabular}{l c c c}
    \toprule
    \textbf{Method} & \textbf{SAcc} & \textbf{Adv$_{\text{PGD}}$} & \textbf{Adv$_{\text{FGSM}}$} \\
    \midrule
    FGSM      & 93.49  & 20.29 & 66.35 \\
    NIFGSM    & 93.31  & 19.77 & 66.84 \\
    VMIFGSM   & 93.32  & 19.15 & 66.85 \\
    PGD       & 93.47  & 19.34 & 66.36 \\
    \bottomrule
    \end{tabular}
    }
\end{minipage}%
\hspace{25pt} 
\begin{minipage}{0.32\textwidth} 
    \centering
    \caption{Comparison of Different Ratios of Adversarial Examples generation applied in simple adversarial training on ResNet-56.}
    \label{table:adversarial_examples_ratios_comparison}
    \setlength{\tabcolsep}{2pt} 
    \resizebox{0.75\textwidth}{!}{
    \begin{tabular}{l c c c}
    \toprule
    \textbf{Ratio} & \textbf{SAcc} & \textbf{Adv$_{\text{PGD}}$} & \textbf{Adv$_{\text{FGSM}}$} \\
    \midrule
    0      & 94.31 & 44.25 & 52.40 \\
    20     & 93.96 & 66.81 & 70.34 \\
    50     & 93.41 & 67.86 & 68.89 \\
    100    & 23.00 & 46.49 & 44.85 \\
    \bottomrule
    \end{tabular}
    }
\end{minipage}
\end{table}

\textbf{Impact of different Adversarial examples:}
Under a FLOPs reduction rate of 64.35\% and an adversarial budget of 8/255, we assessed the influence of various adversarial example generation methods (FGSM~\citep{goodfellow2014explaining}, NIFGSM~\citep{lin2019nesterov}, VMIFGSM~\citep{wang2021enhancing}), and PGD~\citep{madry2017towards}). As shown in \autoref{table:adversarial_examples_comparison}, the differences in robustness and SAcc across these methods are minimal. FGSM showed a slight advantage in robustness, but overall, the variation in adversarial example generation techniques did not significantly impact the final performance, suggesting that the choice of method plays a relatively minor role under these conditions.

\textbf{Impact of different ratios of adversarial examples:}
We investigated the effect of varying the ratio of adversarial examples (from 0\% to 100\%) in ResNet-56 with a FLOPs reduction rate of 64.36\%. The PGD-20 attack was performed with a budget of 2/255, while FGSM used a budget of 8/255.

As shown in \autoref{table:adversarial_examples_ratios_comparison}, increasing the adversarial ratio to 20\% significantly boosts robustness with minimal impact on SAcc. Ratios beyond 50\% provide diminishing returns in robustness and a sharp drop in SAcc, particularly at 100\%, where the accuracy falls drastically to 23\%. This suggests that a moderate adversarial ratio offers an optimal balance between maintaining model accuracy and improving robustness.

\subsection{Comparison of MRPF Results Across Different Datasets}
\begin{table*}[t!]
\caption{Performance of Different Methods and Architectures with FLOPs Reduction for CIFAR-10 and CIFAR-100.}
\label{table: appendix_performance results}
\centering  
\resizebox{0.95\textwidth}{!}{
\begin{tabular}{c c c c c c c c c c}
\toprule
\multirow{2}{*}{Architecture} & \multirow{2}{*}{Method} & \multicolumn{4}{c}{CIFAR-10} & \multicolumn{4}{c}{CIFAR-100} \\
\cmidrule(lr){3-6} \cmidrule(lr){7-10}
 & & FLOPs $\downarrow$ & SAcc & Adv$_{\text{PGD-8}}$ & Adv$_{\text{FGSM}}$ & FLOPs $\downarrow$ & SAcc & Adv$_{\text{PGD-2}}$ & Adv$_{\text{FGSM}}$ \\
\midrule

\multirow{6}{*}{ResNet-18} 
& Dense & 0 & 92.11 & 12.45 & 40.02 & 0 & 75.42 & 16.55 & 18.54 \\
\cmidrule(lr){2-10}
& Taylor & 51.60 & 93.86 & 10.78 & 49.19 & 64.35 & 72.61  & 15.42 & 17.78 \\
& Taylor + MRPF & 55.63 & 93.57 & 20.76 & 66.38 & 69.06 & 74.28 & 33.99 & 35.71 \\
& Magnitude & 51.60 & 93.9 & 10.78 & 48.36 & 64.35 & 72.42 & 15.13 & 17.94 \\
& Magnitude + MRPF & 55.63 & 93.97 & 21.06 & 67.21 & 69.06 & 72.83 & 33.86 & 36.33 \\
\cmidrule(lr){2-10}
& Taylor & 91.05 & 92.19 & 6.41 & 42.64 & 84.18 & 71.34 & 13.78 & 15.06 \\
& Taylor + MRPF & 90.73 & 91.76 & 14.24 & 60.89 & 83.15 & 71.16 & 30.42 & 33.31 \\
& Magnitude & 91.05 & 92.12 & 6.13 & 42.69 & 84.18 & 70.15 & 13.28 & 15.53\\
& Magnitude + MRPF & 90.73 & 91.63 & 15.80 & 62.79 & 83.15 & 70.25 & 27.82 & 31.24 \\

\midrule

\multirow{6}{*}{VGG-16}
& Dense & 0 & 93.76 & 15.28 & 50.87 & 0 & 74.6 & 18.71 & 24.35 \\
\cmidrule(lr){2-10}
& Taylor & 51.28 & 92.01 & 10.11 & 45.84 & 51.27 & 70.85 & 16.09 & 22.39 \\
& Taylor + MRPF & 54.78 & 91.67 & 22.11 & 63.61 & 54.78 & 69.73 & 28.84 & 35.60 \\
& Magnitude & 51.28 & 92.05 & 9.67 & 43.99 & 51.27 & 70.78 & 14.02 & 19.64 \\
& Magnitude + MRPF & 54.78 & 92.22 & 21.32 & 64.48 & 54.21 & 69.73 & 28.69 & 36.12 \\

\bottomrule
\end{tabular}
}
\end{table*}

\begin{table}[t!]
\caption{Performance of Different Fine-tuning Methods on Mobilevit with FLOPs Reduction for CIFAR-10 and CIFAR-100.}
\label{table: performance results-mobilevit}
\centering  
\resizebox{0.95\textwidth}{!}{
\begin{tabular}{c c c c c c c c c c}
\toprule
\multirow{2}{*}{Architecture} & \multirow{2}{*}{Method} & \multicolumn{4}{c}{CIFAR-10} & \multicolumn{4}{c}{CIFAR-100} \\
\cmidrule(lr){3-6} \cmidrule(lr){7-10}
 & & FLOPs $\downarrow$ & SAcc & Adv$_{\text{PGD-8}}$ & Adv$_{\text{FGSM}}$ & FLOPs $\downarrow$ & SAcc & Adv$_{\text{PGD-2}}$ & Adv$_{\text{FGSM}}$ \\
\midrule

\multirow{6}{*}
& Dense & 0 & 93.13 & 21.01 & 52.10 & 0 & 70.69 & 19.03 & 19.88 \\
\cmidrule(lr){2-10}
& Taylor & 64.31 & 91.14 & 16.29 & 43.24 & 51.85 & 68.82 & 16.85 & 17.12 \\
& Taylor + AT & 64.31 & 90.34 & 22.33 & 57.38 & 51.91  & 67.15 & 29.82 & 27.89 \\
& Taylor + MRPF & 60.53 & 91.72 & 26.2 & 62.91 & 59.8  & 67.68 & 29.62 & 27.85\\
& Magnitude & 63.25 & 91.32 & 16.97 & 45.47 & 50.80 & 69.43 & 18.10 & 18.85\\
& Magnitude + AT & 63.25 & 90.4 &22.80 & 55.66 & 50.80 & 68.37 & 30.18 & 28.18\\ {MobileVit-xs}
& Magnitude + MRPF & 59.38 & 91.85 & 27.61 & 62.91 & 59.56  & 67.45 & 29.86 & 27.61\\
\cmidrule(lr){2-10}
& Taylor & 83.63 & 88.09 & 15.10 & 39.80 & 73.83  & 65.42 & 15.59 & 15.21 \\
& Taylor + AT & 83.63 & 87.09 & 18.4 & 50.16 & 73.90  & 62.49 & 24.84 & 23.28 \\
& Taylor + MRPF & 79.95 & 89.13 & 23.54 & 55.73 & 70.76 & 64.52 & 27.73 & 25.54\\
& Magnitude & 82.84 & 89.07 & 14.88 & 37.64 & 73.58  & 65.10 & 16.37 & 15.85\\
& Magnitude + AT & 82.84 & 87.16 & 18.23 & 50.26 & 73.58 & 62.93 & 26.27 &24.82 \\
& Magnitude + MRPF & 79.95 & 89.12 & 23.62 & 54.33 & 70.91  & 65.08 & 27.58 & 25.72\\

\bottomrule
\end{tabular}
}
\end{table}
The results presented in \autoref{table: appendix_performance results}, \autoref{table: performance results-mobilevit}, and \autoref{table: appendix_perf_resnet_tiny_imagenet} illustrate the performance of the Module Robust Pruning and Fine-Tuning (MRPF) framework across CIFAR-10, CIFAR-100, and Tiny-ImageNet datasets. These results highlight MRPF's ability to consistently enhance adversarial robustness while maintaining strong accuracy across various architectures and datasets.

\textbf{CIFAR-10 and CIFAR-100:}  
As shown in \autoref{table: appendix_performance results}, for ResNet-18 and VGG-16, MRPF consistently improved adversarial robustness, especially against FGSM and PGD attacks, compared to baseline Taylor and Magnitude pruning methods. On CIFAR-10, ResNet-18 exhibited a significant improvement in adversarial accuracy under PGD-8 attacks, increasing from 10.78\% (Taylor) to 20.76\% (Taylor + MRPF), representing an improvement of nearly 92.5\%. In CIFAR-100, MRPF delivered an even larger relative increase in adversarial accuracy under PGD-2 attacks, from 15.42\% (Taylor) to 33.99\% (Taylor + MRPF), achieving a 120\% improvement, despite the higher complexity of CIFAR-100. These results suggest that MRPF effectively maintains robustness even under more challenging classification tasks like CIFAR-100.

For VGG-16, MRPF preserved high SAcc (91.67\% in CIFAR-10) while delivering substantial gains in adversarial robustness. For example, under FGSM attacks on CIFAR-10, adversarial accuracy increased from 45.84\% to 63.61\%, marking a 38.7\% improvement. On CIFAR-100, a similar trend was observed with improvements in adversarial accuracy under PGD-2 attacks, highlighting MRPF’s ability to handle both accuracy and robustness effectively in different scenarios.

\textbf{MobileViT-xs:}  
In more complex architectures like MobileViT-xs, MRPF demonstrated even stronger improvements, particularly at higher FLOP reductions, as detailed in \autoref{table: performance results-mobilevit}. For CIFAR-10, adversarial accuracy under PGD-8 attacks improved from 16.97\% (Magnitude pruning) to 27.61\% (Magnitude + MRPF), reflecting a 62.7\% increase. On CIFAR-100, adversarial accuracy under FGSM attacks rose from 18.85\% to 29.86\%, a 58.4\% improvement. These results indicate that MRPF is especially effective in complex architectures like MobileViT-xs, which benefit more from adaptive pruning strategies due to their inherent complexity. This analysis suggests that MRPF not only preserves SAcc but also substantially improves adversarial robustness, particularly in more complex networks like MobileViT-xs. The adaptive nature of MRPF allows it to handle architectures that are more sensitive to pruning, further supporting its broader applicability across diverse network types.

\textbf{Cross-Dataset Comparison and Additional Insights:}  
When comparing the performance of MRPF across datasets, it becomes evident that the framework consistently improves adversarial robustness while maintaining strong SAcc. For larger datasets like Tiny-ImageNet, the improvements in robustness were more substantial, likely due to the increased pruning flexibility afforded by deeper architectures such as ResNet-50. The adversarial accuracy gains under FGSM attacks were particularly pronounced, showing greater improvement in Tiny-ImageNet (182.7\%) compared to CIFAR-10 (67.7\%) or CIFAR-100 (58.4\%).

Moreover, this analysis underscores an interesting observation: both dataset complexity and network architecture influence the efficacy of MRPF. More complex datasets and architectures offer greater opportunities for MRPF to enhance adversarial robustness while maintaining performance. The higher the complexity of the dataset (Tiny-ImageNet $>$ CIFAR-100 $>$ CIFAR-10) and architecture (MobileViT-xs $>$ ResNet-50 $>$ ResNet-18), the more pronounced the improvements in both adversarial and SAcc. This can be attributed to the larger capacity and flexibility of more complex models, which allow MRPF to prune more aggressively while preserving critical features. In challenging datasets like Tiny-ImageNet, MRPF can better focus the model on robustness-related features by pruning less critical neurons, leading to more significant robustness and performance gains.

\subsection{Optimization Algorithm Impact}

\begin{table}[t!]
\caption{Comparison of Different Optimization Algorithms on ResNet-34-Tiny-ImageNet.}
\centering
\resizebox{0.6\columnwidth}{!}{ 
\begin{tabular}{l|ccc|ccc}
\toprule
\multirow{2}{*}{FLOPs $\downarrow$} & \multicolumn{3}{c|}{Magnitude-Step} & \multicolumn{3}{c}{Magnitude-Cosine} \\
\cmidrule(lr){2-7}
 & Acc-avg & Adv$_{\text{PGD-2}}$ & Adv$_{\text{FGSM}}$ & Acc-avg & Adv$_{\text{PGD-2}}$ & Adv$_{\text{FGSM}}$ \\
\midrule
0     & 61.59 & 11.94 & 11.73 & 60.71 & 10.79 & 11.87 \\
19.54 & 62.59 & 13.58 & 13.50 & 61.37 & 11.69 & 12.73 \\
51.51 & 61.93 & 12.67 & 12.76 & 60.48 & 11.01 & 12.57 \\
74.92 & 61.02 & 13.30 & 13.14 & 59.02 & 9.57  & 11.80 \\
\midrule
\multirow{2}{*}{FLOPs $\downarrow$} & \multicolumn{3}{c|}{Random-Step} & \multicolumn{3}{c}{Random-Cosine} \\
\cmidrule(lr){2-7}
 & Acc-avg & Adv$_{\text{PGD-2}}$ & Adv$_{\text{FGSM}}$ & Acc-avg & Adv$_{\text{PGD-2}}$ & Adv$_{\text{FGSM}}$ \\
\midrule
0     & 61.59 & 11.94 & 11.73 & 60.71 & 10.79 & 11.87 \\
19.54 & 62.76 & 11.76 & 12.01 & 61.95 & 10.81 & 12.20 \\
51.51 & 61.69 & 12.14 & 12.08 & 59.95 & 10.42 & 12.10 \\
74.92 & 59.77 & 10.84 & 11.13 & 57.06 & 8.89  & 10.91 \\
\bottomrule
\end{tabular}
}
\end{table}

\begin{figure}[h]
    \centering
    \begin{subfigure}[b]{0.45\textwidth}
        \centering
        \includegraphics[width=\textwidth]{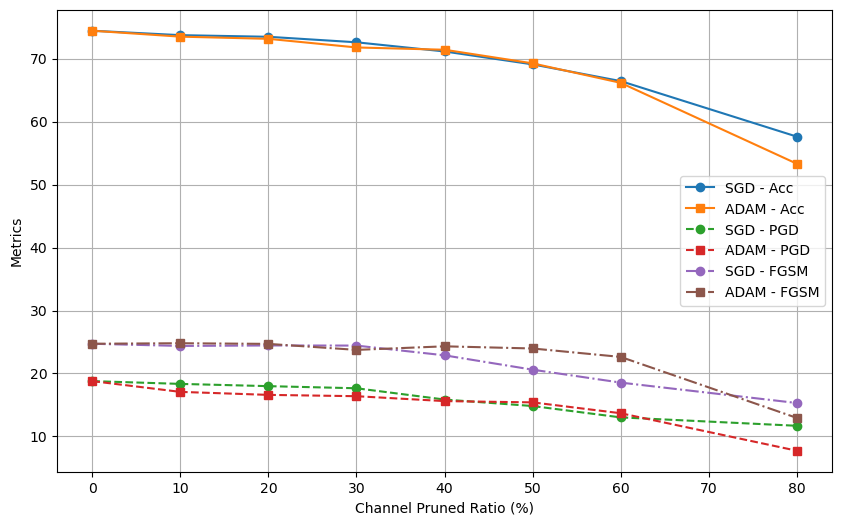}
        \caption{Random Pruning with SGD and ADAM on VGG-16-CIFAR-100}
        \label{fig:random_pruning}
    \end{subfigure}
    \hfill
    \begin{subfigure}[b]{0.45\textwidth}
        \centering
        \includegraphics[width=\textwidth]{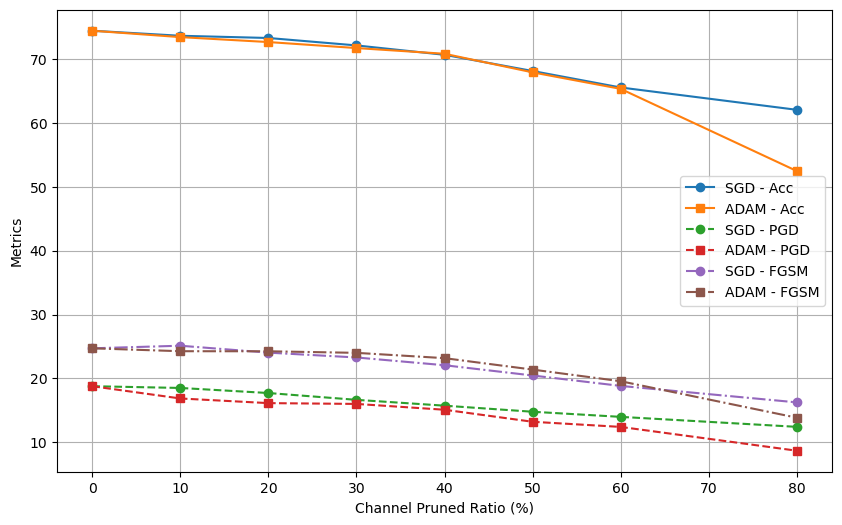}
        \caption{Magnitude Pruning with SGD and ADAM on VGG-16-CIFAR-100}
        \label{fig:magnitude_pruning}
    \end{subfigure}
    \caption{Impact of different optimizers (SGD and ADAM) on pruned models for VGG-16-CIFAR-100 under different pruning strategies. The left figure shows random pruning, and the right figure shows magnitude pruning.}
    \label{fig:pruning_optimizers_comparison}
\end{figure}
The figures \ref{fig:random_pruning} and \ref{fig:magnitude_pruning} highlight the substantial impact of optimizer choice on the performance of pruned models, particularly in the context of adversarial robustness. In both random pruning and magnitude pruning, models optimized with Stochastic Gradient Descent (SGD) consistently outperform those using the ADAM optimizer across a range of channel pruning ratios. This trend is evident not only in terms of SAcc but also in adversarial robustness under PGD and FGSM attacks.

As seen in the figures, as the pruning ratio increases from 0\% to 80\%, the performance gap between SGD and ADAM becomes more pronounced. Specifically, SGD-optimized models maintain higher accuracy and robustness at aggressive pruning levels (e.g., 80\% pruning). For instance, in magnitude pruning, SGD sustains a better adversarial accuracy under PGD attacks, reducing the performance degradation that ADAM suffers at higher pruning levels.

The effectiveness of SGD in preserving model robustness can be attributed to its better generalization capabilities, which are particularly important when the model capacity is reduced through pruning. In contrast, ADAM's adaptive nature may lead to overfitting on smaller, less expressive pruned models, causing a more significant drop in both SAcc and adversarial robustness. These findings suggest that optimizer selection is a critical factor in ensuring that pruned models maintain strong performance, especially under adversarial settings.

\end{document}